\documentclass[aoas]{imsart}

\RequirePackage{amsthm,amsmath,amsfonts,amssymb}
\RequirePackage[authoryear]{natbib}
\RequirePackage[colorlinks,citecolor=blue,urlcolor=blue]{hyperref}
\RequirePackage{graphicx}
 
\startlocaldefs
\theoremstyle{plain}

\newtheorem{theorem}{Theorem}[section]

\theoremstyle{remark}
\newtheorem{definition}{Definition}

\newtheorem{remark}{Remark}
\newcommand{\bx}{\mathbf{x}}

\usepackage{booktabs}
\usepackage{multirow}

\usepackage{algorithmic}
\usepackage{algorithm}

\usepackage{graphicx}
\usepackage{float}
\usepackage{subfig}

\usepackage{amsmath}
\numberwithin{equation}{section}
\usepackage{mathtools}

\endlocaldefs

\begin{document}

\begin{frontmatter}
%\title{Multitask Gaussian Bayesian Network with Application to Personalized Default Mode Network Learning of Major Depressive Disorder Patients}
\title{Learning Multitask Gaussian Bayesian Networks}
%\title{A sample article title with some additional note\thanksref{t1}}
\runtitle{Learning Multitask Gaussian Bayesian Networks}
%\thankstext{T1}{A sample additional note to the title.}

\begin{aug}
%%%%%%%%%%%%%%%%%%%%%%%%%%%%%%%%%%%%%%%%%%%%%%%
%% Only one address is permitted per author. %%
%% Only division, organization and e-mail is %%
%% included in the address.                  %%
%% Additional information can be included in %%
%% the Acknowledgments section if necessary. %%
%%%%%%%%%%%%%%%%%%%%%%%%%%%%%%%%%%%%%%%%%%%%%%%
\author[A]{\fnms{Shuai} \snm{Liu}\ead[label=e1,mark]{hljliushuai@stu.xjtu.edu.cn}},
\author[B]{\fnms{Yixuan} \snm{Qiu}\ead[label=e2]{qiuyixuan@sufe.edu.cn}},
\author[C]{\fnms{Baojuan} \snm{Li}\ead[label=e3]{libjuan@163.com}},
\author[D]{\fnms{Huaning} \snm{Wang}\ead[label=e4]{xskzhu@fmmu.edu.cn}},
\and
\author[A]{\fnms{Xiangyu} \snm{Chang}\ead[label=e5,mark]{xiangyuchang@xjtu.edu.cn}}
%%%%%%%%%%%%%%%%%%%%%%%%%%%%%%%%%%%%%%%%%%%%%%
%% Addresses                                %%
%%%%%%%%%%%%%%%%%%%%%%%%%%%%%%%%%%%%%%%%%%%%%%
\address[A]{School of Management, Xi'an Jiaotong University,
\printead{e1,e5}}

\address[B]{School of Statistics and Management, Shanghai University of Finance and Economics,
\printead{e2}}

\address[C]{School of Biomedical Engineering, Fourth Military Medical University,
\printead{e3}}

\address[D]{Department of Psychiatry, Xijing Hospital, Air Force Medical University, 
\printead{e4}}
\end{aug}

\begin{abstract}
Major depressive disorder (MDD) is closely related to the brain functional connectivity alterations in patients, which can be uncovered by resting-state functional magnetic resonance imaging (rs-fMRI) data. We consider the problem of identifying alterations of the brain functional connectivity for a single MDD patient, which is particularly difficult as the amount of data collected during an fMRI scan is too limited to provide sufficient information for individual analysis. Additionally, rs-fMRI data usually have the characteristics of incompleteness, sparsity, variability, high dimensions, and large noise.

To address the challenges above, especially learning from insufficient rs-fMRI data, we propose a multitask Gaussian Bayesian network framework to identify disorder-induced alterations for individual MDD patients. We achieve this goal by treating each patient in a class of observations as a task, and leveraging data from others. Since the patients have the same type of disorder, they share similar connectivity alterations in brain functional networks. Therefore, we introduce a shared prior covariance matrix among all tasks and utilize the hierarchical model structure to estimate multiple Gaussian Bayesian networks. For computing, we derive closed-form expressions for the complete likelihood function and its gradient and use the Monte Carlo expectation-maximization algorithm to efficiently search for the approximately best network structures. We assess the performance of the proposed method with extensive and systematic numerical experiments and highlight its usefulness in analyzing rs-fMRI data and facilitating the study of MDD.
\end{abstract}


\end{frontmatter}
%%%%%%%%%%%%%%%%%%%%%%%%%%%%%%%%%%%%%%%%%%%%%%
%% Please use \tableofcontents for articles %%
%% with 50 pages and more                   %%
%%%%%%%%%%%%%%%%%%%%%%%%%%%%%%%%%%%%%%%%%%%%%%
%\tableofcontents

\section{Introduction}
\label{sec:introduction}

        Major depressive disorder (MDD) is a common mental disorder characterized by low mood, worthlessness, anxiety, and cognitive impairments \citep{fu2008pattern}. MDD causes more than 800,000 deaths worldwide every year and is the second leading cause of disability worldwide, with point prevalence exceeding 4\% \citep{yan2019reduced}. Therefore, understanding the mechanism of MDD is crucial for effective prevention, diagnosis, and treatment. The existing knowledge about MDD is very limited, but recent studies have found that MDD is closely related to the alterations of brain functional connectivity \citep{wu2011abnormal,liu2013abnormal,yamashita2020generalizable}. Resting-state functional magnetic resonance imaging (rs-fMRI) is an effective modality for understanding the biological and cognitive information of mental disorders \citep{xia2017functional}. It can be used for the noninvasive investigation of MDD by quantifying functional connections of the whole brain that are related to spontaneous blood-oxygen-level-dependent signal fluctuations \citep{smith2013functional}.

Bayesian network (BN), as an approach to studying the cause of interdependence between variables in a system, has drawn much attention \citep{koller2009probabilistic}. BNs visualize the way of information propagation in an intuitive graphical representation \citep{koski2012review}. Such an interpretable structure makes them widely applied in scientific domains such as bioinformatics and neuroscience \citep{bielza2014bayesian}. Many BN models have been developed for disease screening, diagnosis, treatment, and prognosis. For example, \cite{moreira2016preeclampsia} builds a BN model for the diagnosis of preeclampsia, and \cite{zarikas2015bayesian} utilizes the BN model to predict the future severity of acute lung disease.

However, BNs have not been widely used in clinical practice because of the gap between developing an accurate BN and proving its clinical availability. This is more obvious when using BNs to study brain functional connectivity of MDD using rs-fMRI data. The main barriers are: (1) In addition to high dimensionality, rs-fMRI data usually have the characteristics of incompleteness, sparsity, variability, high noise, and significant error. When the number of variables exceeds 50 to 100, the number of possible network structures increases super-exponentially. Although some progress has been made in learning large BNs \citep{gamez2011learning,aragam2017learning}, learning and interpreting BNs from rs-fMRI data is still challenging. (2) Due to the difficult and expensive data acquisition, the amount of rs-fMRI data is relatively small. In the Strategic Research Program for Brain Sciences database, only 255 MDD patients' rs-fMRI data were collected \citep{yamashita2021common}. Moreover, because of the considerable heterogeneity of depression, the brain functional connectivity in patients with MDD varies from person to person \citep{young2016anhedonia}, and there are different subtypes of depression itself \citep{beijers2019data}. Therefore, the available rs-fMRI datasets are too small to simultaneously learn robust BNs for individuals or subtype groups of MDD. Learning more helpful information from limited data is still one of the research hotspots of BN learning.

To learn robust BNs efficiently via insufficient rs-fMRI data, we propose a multitask learning (MTL) method to learn multiple Gaussian Bayesian networks (GBNs) simultaneously, which we call the multitask Gaussian Bayesian network (MTGBN). Our original interest in developing this method arises from the noninvasive biomarker discovery in rs-fMRI data. For example, in the study of brain functional connectivity alterations for a single MDD patient, data derived from an individual's neuroimage do not deliver enough knowledge due to the complex data structure and high data dimensionality. However, since the patients are related in the sense that they have the same disorder, they share disorder-induced connectivity alteration in brain functional networks. Therefore, in the BN learning, we also hope to leverage data from other patients with the assumption that such alterations show some degrees of similarity. Such models would help practitioners understand how systems are structured jointly from related tasks.

Few works exist on using fMRI data to learn network structures from related tasks to our best knowledge. \cite{chen2021joint} considers estimating multiple differential networks with fMRI datasets from multiple research centers and exploits sparse group minimax concave penalty to induce sparse structures of each differential network. \cite{he2021transfer} proposes an algorithm called Trans-Copula-CLIME for estimating undirected graphical models from fMRI datasets, which characterizes the similarity between tasks by the sparsity of a divergence matrix. Different from the work by \cite{chen2021joint} and \cite{he2021transfer}, we focus on GBN learning. The benefit that GBN brings to studying network structures is that GBN can characterize the direct dependencies between variables, which can provide more information than differential networks and undirected graphical models. This is the first work to use the Bayesian approach to learn multiple GBNs from related tasks on MTL, as far as we know.

The proposed MTGBN method improves the learning ability of GBNs via the MTL technique. Specifically, we treat each subject in a class of observations as a task and assume that the tasks share some common prior knowledge. A Bayesian hierarchical model thus models the observed data. This setting can help us learn useful information from limited data. Remarkably, in the structure learning of BNs, we derive a closed-form formula for the complete likelihood function and its gradient and use the
Monte Carlo expectation-maximization algorithm (MCEM, \citealp{wei1990monte}) to efficiently search for the approximately best BN structures. The contributions of this article are threefold. 
\begin{itemize}
    \item The proposed MTGBN method is able to learn BNs from insufficient high dimensional data by jointly learning multiple BN structures from related observations. It addresses one common problem in pattern recognition: how to learn network structures from related observations such that the representations can share related information while also maintaining subject-specific characteristics.
    \item MTGBN overcomes the computational intractability issue of the existing MTL framework for GBNs. We derive a closed-form expression for the complete likelihood function and provide exact gradient evaluations to the Hamiltonian Monte Carlo (HMC, \citealp{neal2011mcmc}) sampler that is used by the MCEM algorithm.
    \item We apply MTGBN to rs-fMRI data of MDD patients and show that MTGBN can effectively identify the increase of functional connectivity in MDD patients, especially the connectivity within the dorsal medial prefrontal cortex. Doctors can use it to understand the potential abnormalities of brain functional connections of individual MDD patients and carry out personalized treatments.
\end{itemize}

This article is organized as follows. In Section \ref{sec:data} we begin by introducing the rs-fMRI dataset analyzed in this work. In Section \ref{sec:background}, we briefly review BN learning and MTL methods. We present the formulation of MTGBN in Section \ref{sec:MTGBN}, and introduce its structure learning in Section \ref{sec:structure_learning_mtgbn}. An efficient computing algorithm for MTGBN is given in Section \ref{sec:computing}. We conduct simulations in Section \ref{sec:simulation}, and analyze a real rs-fMRI dataset and assess the advantages of MTGBN in Section \ref{sec:applications}. Finally, we conclude this article in Section \ref{sec:conclusion}.

\section{Real-world rs-fMRI Data}
\label{sec:data}

\subsection{Data acquisition}
\label{subsec:acquisition}

The data analyzed in this study were obtained from the DecNef Project Brain Data Repository\footnote{\url{https://bicr-resource.atr.jp/srpbsopen/}}, and are a subset of the SRPBS multi-disorder MRI dataset \citep{tanaka2021multi}. The rs-fMRI data for each participant at resting state were obtained using 3-Tesla Siemens whole body MRI systems at Center of Innovation in Hiroshima University. During the fMRI scan, participants were instructed to be relax, stay awake and refrain from thinking anything particular. The rs-fMRI data have the following parameters: TR = 2.5s, TE = 30ms, flip angle = $80^\circ$, matrix size = $64\times 64$, field of view = $212$mm, in-plane resolution = $3.3\times 3.3$, slice gap = 0.8mm, slice thickness = 3.2mm, number of slices = 40, number of volumes = 240 + 4 (dummy), total scan time = 10min + 10s (dummy). The T1-weighted images were then acquired with the following parameters: voxel size = $1\times 1\times 1$, TR = 2300ms, TE = 2.98ms, TI = 900ms; flip angle = $90^\circ$, FOV = $256$, matrix = $256\times 256$.

\subsection{Samples}
\label{subsec:samples}

We use the rs-fMRI data of 15 MDD patients and 15 healthy control (HC) participants for the experiment. The selection criteria are as follows. The MDD patients are aged between 40 and 50, right-handed, free of other diseases, and their BDI scores are between 20 and 35. BDI is a clinical instrument for the severity of depression. The higher the score, the more severe the depression. The HC participants are selected to purposely match the MDD patients in terms of other selection criteria except for BDI score, \emph{i.e.}, they are in the same age cohort as the MDD patients, right-handed, free of other diseases, but with a BDI score between 0 and 7 (\emph{i.e.}, they do not have depression).

We applied the following pre-processing steps to the data. For each participant, 230 fMRI volumes were remained after removing the first ten volumes. Slice acquisition time and head motion were corrected on the remaining images. Participants with excessive head motion were not enrolled. The T1-weighted images use unified segment and diffeomorphic anatomical registration to guide fMRI registration through exponentiated Lie algebra as implemented in the SPM12 software\footnote{\url{https://www.fil.ion.ucl.ac.uk/spm/software/spm12/}}. The rs-fMRI data were smoothed with a 6mm full-width at half-maximum Gaussian kernel and further denoised by regressing out several nuisance signals, including the Friston-24 head motion parameters and signals from cerebrospinal fluid and white matter, before linear detrending and temporal band-pass filtering ($0.01-0.08$ Hz).

After pre-processing, we select 30 brain regions (each viewed as a region of interest, ROI) based on three different network templates related to the default mode network (DMN, \citealp{raichle2007default}), including the Andrews-Hanna DMN (aDMN, \citealp{buckner2008brain}), the dorsal DMN (dDMN, \citealp{chen2017dissociated}), and the ventral DMN (vDMN, \citealp{lee2021ventral}). Details of brain regions are presented in Table \ref{tab: roi}. The voxel-wise fMRI time courses are averaged into one regional average time course within each selected region. Then each participant's dataset has the form of a $230\times 30$ matrix (230 sampling points in each regional average fMRI time course and 30 selected brain regions). The dataset obtained through the steps above can be reasonably considered to follow Gaussian distributions because each brain region includes at least several hundred voxels and the measurement data of each region is the regional average over the belonging voxels. Therefore, the central limit theorem is applicable. Also, such hypotheses are common for fMRI studies \citep{valdes2011effective}.

\section{Background and related work}
\label{sec:background}

\subsection{Learning Bayesian networks}
\label{subsec:BNs}

BN is a formal method to analyze decision-making strategies under uncertain conditions \citep{pearl1988probabilistic}.
Given a random vector $\boldsymbol{X}=(X_{1}, \ldots, X_{p})^\top$, we use $\mathbb{P}(\bx)$ to denote its joint probability mass function (for the discrete case) or the joint density function (for the continuous case), where $\bx$ is an observation of $\boldsymbol{X}$. From the chain rule of probability, we can write $\mathbb{P}(\bx)$ as
\begin{equation}
\mathbb{P}(\bx)=\mathbb{P}(x_{1})\prod_{i=2}^{p} \mathbb{P}\left(x_{i} \mid x_{1}, \ldots, x_{i-1}\right).
\label{eq:Px}
\end{equation}
For each $X_{i}$, let $\Pi_{i} \subseteq\left\{X_{1}, \ldots, X_{i-1}\right\}$ be a set of variables such that
\begin{equation}
\mathbb{P}\left(x_{i} \mid x_{1}, \ldots, x_{i-1}\right)=\mathbb{P}\left(x_{i} \mid \Pi_{i}\right).
\label{eq:Pxcond}
\end{equation}

A BN $B=(\mathcal{G}, \Theta)$ can represent the joint distribution of $\boldsymbol{X}$ through a directed network model, where $\mathcal{G}$ is a network structure that encodes the assertions of conditional independence in equation \eqref{eq:Pxcond}, and $\Theta$ includes the parameters of the BN model corresponding to $\mathcal{G}$ \citep{koller2009probabilistic}. In particular, $\mathcal{G}$ is a directed acyclic graph (DAG) such that (1) each variable in $\boldsymbol{X}$ corresponds to a node in $\mathcal{G}$, and (2) the parents of the node corresponding to $X_{i}$ are the nodes corresponding to the variables in $\Pi_{i}$. Throughout this article, we use $X_{i}$ to refer to both the variable and its corresponding node in the graph. Combining equations \eqref{eq:Px} and \eqref{eq:Pxcond}, we can see that the joint distribution for $\boldsymbol{X}$ can be represented graphically by a BN, namely,
\[
\mathbb{P}(\bx)=\mathbb{P}(x_{1})\prod_{i=2}^{p} \mathbb{P}\left(x_{i} \mid \Pi_{i}\right).
\]

GBNs are a class of important and widely-used BNs \citep{geiger1994learning}. In GBNs, $\boldsymbol{X}$ follows a multivariate normal distribution $\mathcal{N}\left(\mu, \Sigma \right)$, with the joint density function
\[
   \mathbb{P}(\bx)= (2 \pi)^{-p/2}\left|\Sigma\right|^{-1/2}\exp\left(-\frac{1}{2}(\bx-\mu)^{\top} \Sigma^{-1}(\bx-\mu)\right),
\]
where $\mu$ is the expectation of $\boldsymbol{X}$, $\Sigma$ is the covariance matrix, and $|\Sigma|$ is the determinant of $\Sigma$. Clearly, the parameters of a GBN are $\Theta=\{\mu,\Sigma\}$.

The structure $\mathcal{G}$ and the parameters $\Theta$ can be learned from the observed data, which we call the BN learning procedure.
The primary purpose of BN learning is to estimate a BN $B=(\mathcal{G}, \Theta)$ that can best explain the $n$ independent observations $D=\{\bx_{i}=(x_{i1},\dots,x_{ip})^{\top}\}_{i=1}^{n}$, leading to two major tasks: the structure learning, which attempts to learn a DAG $\mathcal{G}$ that encodes the structural dependence of the variables in $\boldsymbol{X}$, and the parameter learning, which estimates the parameters $\Theta$ associated with $\mathcal{G}$.

\paragraph*{Structure learning}

Structure learning is an essential part of BNs, which can mainly be divided into two categories: (1) constraint-based methods; (2) score-based methods. Constraint-based methods try to identify dependent and independent areas within the data \citep{koller2009probabilistic}. One widely-used method for structure learning is the SGS algorithm \citep{spirtes2000causation} that identifies the dependencies of variable pairs by evaluating their independence. This method is computationally expensive because the number of tests overgrows as the network size increases. Compared with SGS, the PC algorithm \citep{spirtes1991algorithm} is more efficient because it has a size threshold. The PC algorithm assumes that all nodes in the undirected graph are connected and uses hypothesis tests to delete the connections.

Score-based methods attempt to compute a score measuring all possible network structures, using a scoring function such as BIC \citep{schwarz1978estimating}, K2 \citep{cooper1992bayesian}, and BDeu \citep{heckerman1995learning}. Because learning BN structures from data is an NP-hard problem, existing score-based methods usually use heuristic search techniques to reduce the computational cost. For large networks, the search space is extremely large, and ordinary score-based methods also become intractable.
Therefore, \cite{friedman2013learning} proposes to improve the learning efficiency of score-based methods by restricting the search space.

\paragraph*{Parameter learning}

There are two main methods for parameter learning: (1) maximum likelihood estimation (MLE, see \citealp{ji2015review}); (2) Bayesian estimation \citep{geiger1997characterization}. MLE method estimates the parameters by maximizing the likelihood of the data. However, MLE becomes problematic when the data are sparse, as the dataset does not contain all possible combinations of the involved variables. In such cases, the Bayesian estimation is preferred, which places a prior distribution on the variables \citep{altendorf2005learning}. It should be pointed out that parameter learning itself is not sufficient, as it is based on the assumption that the network structure is known in advance, or at least one structure is held fixed during parameter learning.\smallskip{}

In this work, we focus on the score-based structure learning. The score functions for BN are typically derived via a Bayesian approach. For example, one Bayesian measure of the goodness of a BN structure $\mathcal{G}$ is its posterior probability given the data $D$:
\begin{equation}
\mathbb{P}(\mathcal{G} \mid D) = \frac{1}{c}\ \mathbb{P}(\mathcal{G}) \times \mathbb{P}(D \mid \mathcal{G}),
\label{eq:Pgd}
\end{equation}
where $\mathbb{P}(\mathcal{G})$ is the prior of  $\mathcal{G}$, and $c=\mathbb{P}(D)=\sum_{\mathcal{G}} \mathbb{P}(\mathcal{G}) \times \mathbb{P}(D \mid \mathcal{G})$ is a normalizing constant. Computing $c$ is in general intractable, as there may be a huge number of network structures to sum over, even in a small domain. But we never need to explicitly compute $c$, since maximizing the posterior probability is equivalent to maximizing the joint distribution $\mathbb{P}(\mathcal{G}) \times \mathbb{P}(D \mid \mathcal{G})=\mathbb{P}(D ,\mathcal{G})$, which we use as the score function in what follows.

Once the score function is determined, it can then be used to find the best $\mathcal{G}$ among the set of all network structures. Unfortunately, this optimization problem is NP-hard. Therefore, heuristic algorithms, including greedy search, greedy search with restarts, best-first search, and Monte Carlo methods, are usually required \citep{heckerman1995learning}.

\subsection{Multitask learning}
\label{sebsec:MTL}

General machine learning methods usually require many samples to learn a robust model. In the case of insufficient data, MTL \citep{caruana1997multitask} is a general approach that incorporates task relatedness during learning and can learn robust models \citep{zhang2021survey}. Given $m$ learning tasks $\mathcal{T}=\left\{T_{1}, \ldots, T_{m}\right\}$, where at least one of them is related, the aim of the MTL is to improve the performance and generalization of a model $T_{i}$ by using the knowledge contained in all or some of related $m$ tasks. The tasks in MTL can be general learning tasks such as supervised tasks \citep{schwab2018not}, unsupervised tasks \citep{cao2019unsupervised}, semi-supervised tasks \citep{rei2017semi}, reinforcement learning tasks \citep{teh2017distral}, or graphical tasks \citep{honorio2010multi}.

According to \cite{zhang2021survey}, the current research of MTL focuses on characterizing the relatedness between tasks, which can be divided into three classes, including instance-based MTL \citep{bickel2008multi}, feature-based MTL \citep{evgeniou2007multi}, and parameter-based MTL \citep{parameswaran2010large}. In this paper, we focus on the parameter-based MTL. The most common approach in parameter-based MTL is the task relation learning approach, which quantifies the relatedness among tasks in three ways: task similarity, task correlation, and task covariance. Early work such as \cite{evgeniou2004regularized} considers the task relations as prior information. Several studies have taken the relatedness into account by placing a common prior on model parameters of the tasks \citep{evgeniou2005learning}, e.g., with Gaussian copula models \citep{gonccalves2016multi} and matrix generalized inverse Gaussian distribution \citep{li2015bayesian}. However, when task relations are not available in applications, it is necessary to automatically learn task relationships from data. For instance, \cite{williams2007multi} proposes a multitask Gaussian process to learn task covariance from data, and \cite{zhang2010multi} proposes a multitask generalized $t$ process by placing an inverse-Wishart prior.

\section{The proposed MTGBN framework}
\label{sec:MTGBN}

The MTGBN framework is primarily motivated by various real-world problems. Consider a medical task that studies a specific disease with $m$ subtypes, and suppose that the collected variables for each subtype are modeled by a BN. However, the data sizes for some subtypes are small, making it challenging to learn robust network structures for all subtypes. Also, it is inappropriate to pool the data and only learn a single BN shared by all the subtypes, as it ignores the heterogeneity between subtypes.

In the above scenario, the MTL method that learns multiple BNs simultaneously is much preferred, where we treat the modeling of each subtype as one task. In this article, we focus on GBNs and propose the MTGBN model that can address the issue of insufficient sample sizes in individual GBN learning tasks. MTGBN achieves this target by assuming that the $m$ subtypes share some common information, compensating for the lack of data for minor subtypes.

Formally, let $\mathcal{D}=\left\{D_{1}, \ldots, D_{m}\right\}$ be $m$ related observational datasets, where $D_{i}=\{\bx_{i1},\ldots,\bx_{in_{i}}\}$ represents the collected data for task $i$. Assume that each observation $\bx_{ij}\in\mathbb{R}^p$ is the realization of a multivariate normal distribution with zero mean and covariance matrix $\Sigma_{i}$.

\begin{figure}
\includegraphics[width=0.7\textwidth]{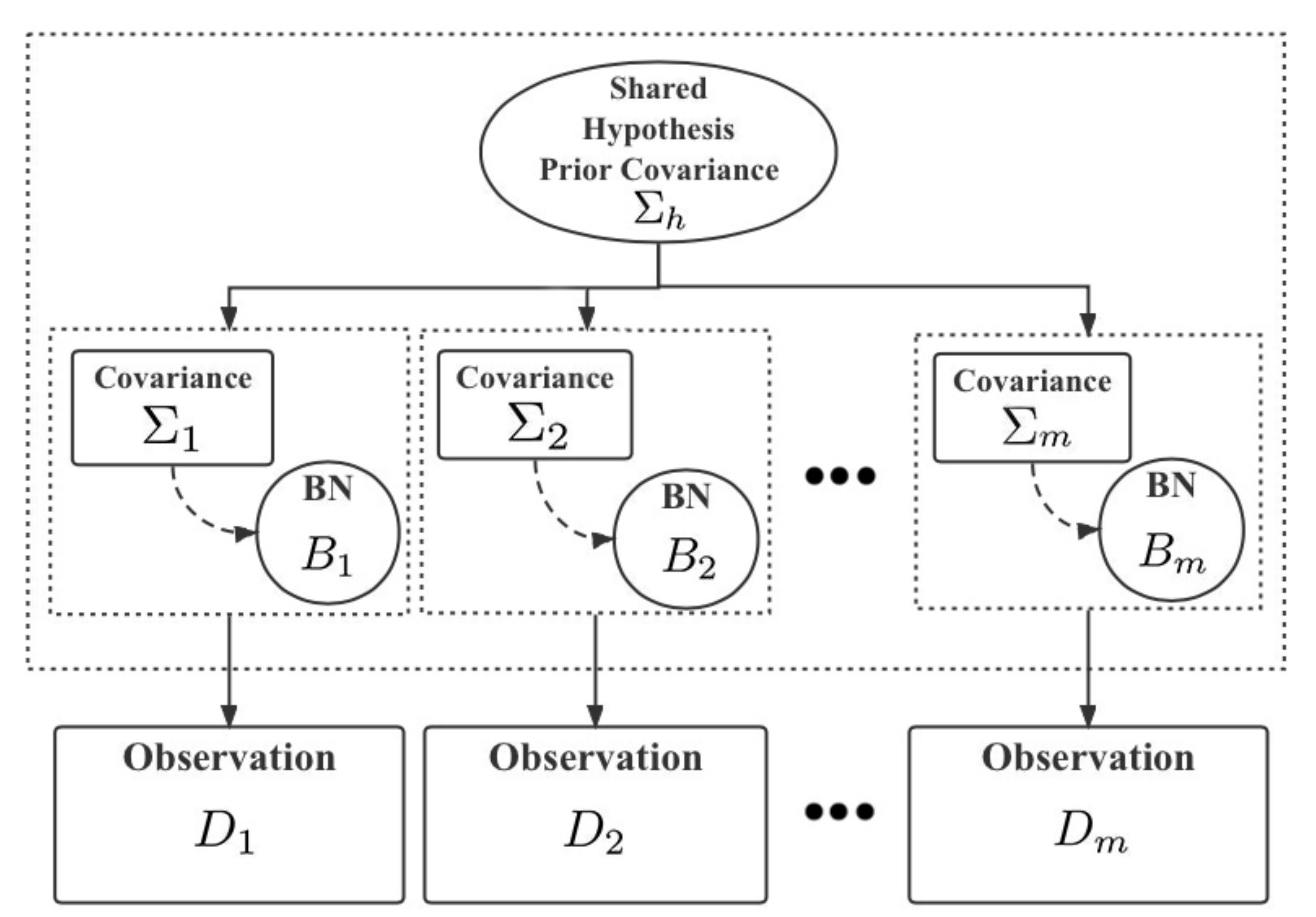}
\caption{Illustration of the generative framework of MTGBN}
\label{framework}
\end{figure}

In many applications, we have prior beliefs about the relatedness of tasks based on metadata or domain knowledge. In MTGBN, we define a shared prior parameter, $\Sigma_h$, to represent the relatedness of all tasks. Figure \ref{framework}  shows the generative framework of the MTGBN method, where the covariance matrices $\left\{\Sigma_{i}\right\}_{i=1}^{m}$ are sampled from a common prior distribution parameterized by $\Sigma_h$, and each $\Sigma_i$ generates a dataset $D_i$ for task $i$. $\Sigma_h$ is inferred from all the tasks, which helps transfer the learned knowledge from one task to another.

We formalize the proposed MTGBN framework as follows. First, the parameter $\Sigma_h$ is assigned a uniform prior distribution over all $p\times p$ positive definite matrices. Then given $\Sigma_h$, the $m$ covariance matrices $\boldsymbol{\Sigma}=\left\{\Sigma_{i}\right\}_{i=1}^{m}$ are independently sampled from an inverse Wishart distribution $I W\left(\Sigma_{h}, \nu_{0}\right)$ \citep{robert2014machine}, where $\nu_{0}>p-1$ and with the following joint density function:
\begin{equation}
\begin{aligned}
&\mathbb{P}\left(\boldsymbol{\Sigma} \mid \Sigma_{h}\right)=\prod_{i=1}^{m} \mathbb{P}\left(\Sigma_{i} \mid \Sigma_{h}\right),\\
&\mathbb{P}\left(\Sigma_{i} \mid \Sigma_{h}\right)=\frac{\left|\Sigma_{h}\right|^{\nu_{0} / 2}}{2^{\nu_{0} p / 2} \Gamma_{p}\left(\nu_{0} / 2\right)}\left|\Sigma_{i}\right|^{-\left(\nu_{0}+p+1\right) / 2}\cdot\exp \left\{-\frac{1}{2} \operatorname{tr}\left(\Sigma_{h} \Sigma_{i}^{-1}\right)\right\},
\label{eq:Psigma}
\end{aligned}
\end{equation}
where for $\alpha>(m-1)/2$,
\[
\Gamma_{m}(\alpha)=\pi^{m(m-1) / 4} \prod_{i=1}^{m} \Gamma\left(\alpha-(i-1)/2\right)
\]
is the multivariate gamma function \citep{muirhead2009aspects}, $\Gamma(\cdot)$ is the gamma function, $p$ denotes the dimension of $\Sigma_{i}$, and $\nu_{0}$ is the degree of freedom in the inverse Wishart distribution, viewed as a fixed hyperparameter.

Finally, we assume that the observed data $D_{i}$ in class $i$ follow a zero mean multivariate normal distribution, and the $m$ datasets $\mathcal{D}=\left\{D_{i}\right\}_{i=1}^{m}$ are mutually independent given the covariance matrices $\boldsymbol{\Sigma}=\left\{\Sigma_{i}\right\}_{i=1}^{m}$. Clearly, the joint density function of  $\mathcal{D}$ given $\boldsymbol{\Sigma}$ is
\begin{equation}
\begin{aligned}
&\mathbb{P}\left(\mathcal{D} \mid\boldsymbol{\Sigma}\right)=\prod_{i=1}^{m} \mathbb{P}\left(D_{i} \mid \Sigma_{i}\right),\\
&\mathbb{P}\left(D_{i} \mid \Sigma_{i}\right)=(2 \pi)^{-n_{i} p / 2}\left|\Sigma_{i}\right|^{-n_{i} / 2} \exp \left\{-\frac{1}{2} \operatorname{tr}\left(n_{i} \Sigma_{i}^{-1} S_{i}\right)\right\},
\end{aligned}
\label{eq:Pdsigmai}
\end{equation}
where $n_i$ denotes the sample size of $D_{i}$, and $S_{i}$ is the sample covariance matrix of $D_{i}$.

The parameter learning of MTGBN is concerned with inferring all the parameters $\Sigma_h $ and $\boldsymbol{\Sigma}$ from $\mathcal{D}=\left\{D_{1}, \ldots, D_{m}\right\}$. However, one major advantage of GBN is that it can reveal the essential causal relationship between target variables, and hence in this work our main interest is on the structure learning of MTGBN. For convenience, in what follows we omit the parameter component $\Theta$ in a BN $B=(\mathcal{G}, \Theta)$, and use $B_i$ to represent the BN structure of the collected dataset for class $i$. Therefore, the learning goal of MTGBN is to simultaneously infer the $m$ GBNs $\mathcal{B}=\left\{B_{1}, \ldots, B_{m}\right\}$ from $m$ related datasets $\mathcal{D}=\left\{D_{1}, \ldots, D_{m}\right\}$, with $B_i$ corresponding to $D_i$.

\section{Structure learning of MTGBN}
\label{sec:structure_learning_mtgbn}

As is described in Section \ref{subsec:BNs}, to learn the BN structures from observations, we need to first construct an appropriate score function, and then use heuristic methods to search for the best network structure based on the score function. Similar to equation \eqref{eq:Pgd}, we follow the likelihood principle, \emph{i.e.}, taking $\mathbb{P}\left(\mathcal{B}, \mathcal{D}\right)$ as the score function. However, one major computational challenge of MTGBN stems from the hypothesis prior covariance matrix $\Sigma_{h}$. Although the introduction of $\Sigma_{h}$ allows the sharing of information among tasks, it also brings difficulties to maximizing the score function. The hierarchical model illustrated by Figure \ref{framework} defines the joint distribution of $\Sigma_{h}$, $\mathcal{B}$, and $\mathcal{D}$, and we treat the parameter $\Sigma_{h}$ as a latent variable. Therefore, $\mathbb{P}\left(\mathcal{B}, \mathcal{D}\right)$ is computed by integrating out $\Sigma_{h}$ in the joint density function $\mathbb{P}\left(\Sigma_{h}, \mathcal{B}, \mathcal{D}\right)$,
\begin{equation}
\mathbb{P}\left(\mathcal{B}, \mathcal{D}\right)=\int \mathbb{P}\left(\Sigma_{h}, \mathcal{B}, \mathcal{D}\right) \mathrm{d} \Sigma_{h},
\label{eq:Pgdb}
\end{equation}
and we typically work on the log-likelihood $\mathcal{L}\left(\mathcal{B},\mathcal{D}\right)=\log \mathbb{P}\left(\mathcal{B}, \mathcal{D}\right)$ for convenience.

Due to the existence of the latent variable $\Sigma_{h}$, direct evaluation and maximization of $\mathcal{L}\left(\mathcal{B},\mathcal{D}\right)$ is intractable. Fortunately, the expectation-maximization (EM, \citealp{dempster1977maximum}) algorithm provides an effective way to tackle this problem. Specifically, we treat the latent variable $\Sigma_{h}$ as missing data, and the EM algorithm proceeds by utilizing the following two ingredients: (1) the complete log-likelihood function $\log\mathbb{P}\left(\Sigma_{h}, \mathcal{B}, \mathcal{D}\right)$; (2) expectations with respect to the conditional distribution $\mathbb{P}\left(\Sigma_{h} \mid \mathcal{B}, \mathcal{D}\right)$. We first derive the closed-form expressions for these two quantities in Sections \ref{subsec:complete_likelihood} and \ref{subsec:conditional_distr}, respectively, and then outline the EM algorithm in Section \ref{subsec:em_algorithm}.

\subsection{The complete likelihood}
\label{subsec:complete_likelihood}

 The complete likelihood admits the decomposition
\begin{equation}
\mathbb{P}\left(\Sigma_{h}, \mathcal{B}, \mathcal{D}\right)=\mathbb{P}\left(\Sigma_{h}\right)\prod_{i=1}^{m}\mathbb{P}\left(B_{i} \mid \Sigma_{h}\right)\mathbb{P}\left(D_{i} \mid B_{i}, \Sigma_{h}\right).
\label{eq:completell}
\end{equation}

In Section \ref{sec:MTGBN} we have assumed that $\mathbb{P}\left(\Sigma_{h}\right)$ follows a uniform distribution, and to show the expressions for the remaining two parts, $\mathbb{P}\left(B_{i} \mid \Sigma_{h}\right)$ and $\mathbb{P}\left(D_{i} \mid B_{i}, \Sigma_{h}\right)$, it is necessary to introduce a number of definitions and notations.

\begin{definition}
Let $G=(V,E)$ be an undirected graph, where $V=\{1, \dots, d\}$ is the set of nodes and $E$ is the set of undirected edges. Any subset of edges $E^{\prime} \subset E$ can be used to define a \emph{subgraph} $G^{\prime}=(V,E^{\prime})$. A subgraph is called \emph{complete} if every pair of nodes are adjacent. Define a \emph{clique} of $G$ as a maximal complete subgrah of $G$. By maximal, we mean that a clique is not contained in a lager complete subgraph. Define a \emph{decompsoition} of $G=(V,E)$ as a pair $(A,B)$ of subsets of $V$ such that $V=A \cup B$ and $A \cap B$ is complete. An undirected graph is \emph{decomposable} if it is complete, or if there exists a proper decomposition $(A,B)$ into complete subgraphs $G^{A}$ and $G^{B}$ \citep{dawid1993hyper}. In other words, a decomposable graph can be decomposed into several cliques, and each clique is a complete subgraph of the graph.
\end{definition}

Let $\widetilde{G}$ be a decomposable undirected graph consisting of $k$ cliques in the sequence $\left\{C_{1}, \ldots, C_{k}\right\}$. Define the \emph{histories} of the sequence as the sets $H_{j}=\bigcup_{i=1}^{j} C_{i}$ for $j=1, \ldots, k$. Similarly, define the \emph{separators} of the sequence as $S_{j}= H_{j-1}\cap C_{j}$ for $j=2, \ldots, k$. The merit of this decomposition is that we can find the corresponding low-dimensional factorization for a high-dimensional graph. For any two subsets $A$ and $B$, $S$ is said to separate $A$ and $B$, denoted by $A \perp B \mid S$, if for every node pair $a \in A$ and $b \in B$, each path connecting $a$ and $b$ includes at least one element of $S$. A probability density $\mathbb{P}$ is said to be globally Markov, if $X=(X_{A},X_{B}, X_{D})\sim \mathbb{P}$ and $X_{A} \perp X_{B} \mid X_{D}$ for any triple subsets $A$, $B$, $D$ such that $A \perp B \mid D$. $X_{A} \perp X_{B} \mid X_{D}$ means that $X_{A}$ and $X_{B}$ are independent conditional on $X_{D}$. Let $\mathbb{P}_{V}$ be a joint density that is globally Markov with respect to a decomposable graph $\widetilde{G}=(V,E)$.  Let $\mathcal{C}$ and $\mathcal{S}$ be the cliques and separators of $\widetilde{G}$, with corresponding marginal densities $\left\{\mathbb{P}_{C}: C \in \mathcal{C}\right\}$ and $\left\{\mathbb{P}_{S}: S \in \mathcal{S}\right\}$. Then $\mathbb{P}_{V}$ can be factorized as \citep{lauritzen1996graphical}
\[
\mathbb{P}_{V}\left(x_{V}\right)=\frac{\prod_{C \in \mathcal{C}} \mathbb{P}_{C}\left(x_{C}\right)}{\prod_{S \in \mathcal{S}} \mathbb{P}_{S}\left(x_{S}\right)}.
\]

\begin{figure}
\includegraphics[width=0.7\textwidth]{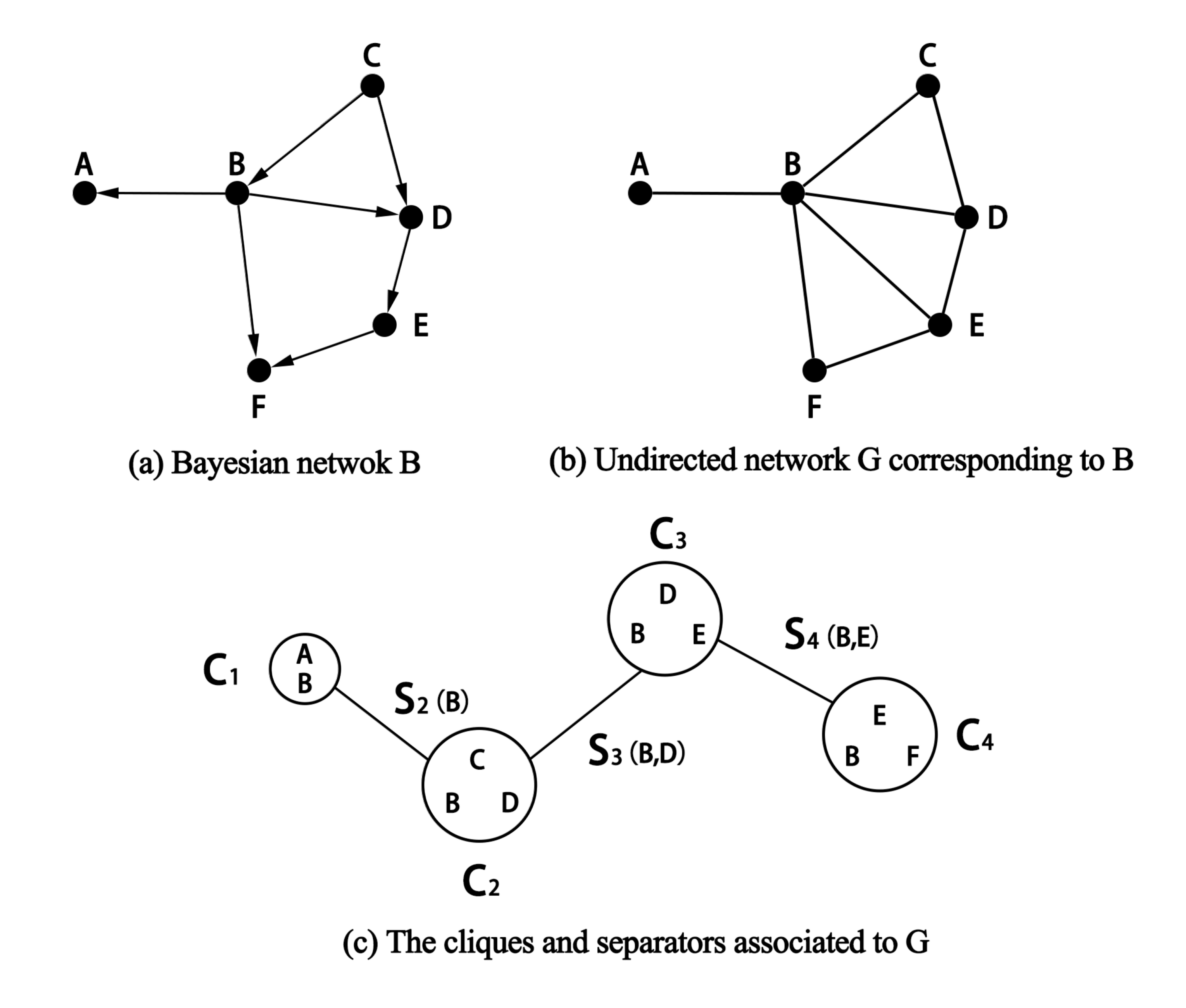}
\caption{An example of graph tranformation and decomposition.}
\label{fig:graph_trans}
\end{figure}

A BN $B$ can be transformed into a unique undirected graph $G$, also known as a moral graph. The undirected graph $G$ is generated from $B$, where if a node has more than one parent node, edges between these parent nodes are inserted, and all directed edges become undirected. As an illustration, Figure \ref{fig:graph_trans}(a) and (b) show the transformation from a BN to an undirected graph, which is achieved by (1) adding an edge between the nodes $V_{B}$ and $V_{E}$, as both $V_{B}$ and $V_{E}$ are the parent nodes of $V_{F}$, and (2) deleting all the directions in $B$. Additionally, Figure~\ref{fig:graph_trans}(c) demonstrates that $G$ can be decomposed into four cliques and three separators.

For Gaussian graphical models, the graph structure is usually related to the inverse of the covariance matrix. Let $M$ be a $p \times p$ positive definite matrix and $A$ is an index set with $A \subset \{1,\dots,p\}$. Then we use $M_{AA}$ to denote the corresponding $|A| \times |A|$ submatrix of $M$, which is also a positive definite matrix. We are now ready to present the first main theorem based on the definitions above.

\begin{theorem}
\label{Theorem1}
Suppose that $\tilde{G}_i$ is the decomposable undirected graph generated by $B_i$, and $\mathcal{C}$ and $\mathcal{S}$ are the cliques and separators of $\widetilde{G}_{i}$, respectively. Then $\mathbb{P}\left(B_{i} \mid \Sigma_{h}\right)$ has the closed-form expression
\begin{equation}
\mathbb{P}\left(B_{i} \mid \Sigma_{h}\right)=\frac{\prod_{C \in \mathcal{C}}\left|2^{-1} \Sigma_{h C C}\right|^{\left(\nu_{0}+|C|-1\right) / 2} \Gamma_{|C|}^{-1}\left(\frac{\nu_{0}+|C|-1}{2}\right)}{\prod_{S \in \mathcal{S}}\left|2^{-1} \Sigma_{h S S}\right|^{\left(\nu_{0}+|S|-1\right) / 2} \Gamma_{|S|}^{-1}\left(\frac{\nu_{0}+|S|-1}{2}\right)}\coloneqq h(\Sigma_{h},\widetilde{G}_{i}),
\label{eq:hfunc}
\end{equation}
where $|C|$ and $|S|$ are the cardinality of the clique $C$ and the separator $S$, respectively, $\Sigma_{h C C}$ and $\Sigma_{h S S}$ denote the submatrices of $\Sigma_{h}$ according to the graph decomposation.
\end{theorem}

We make the following remark to further explain the implications of Theorem \ref{Theorem1}.

\begin{remark}
$\mathbb{P}\left(B_{i} \mid \Sigma_{h}\right)$ can be expressed as $\mathbb{P}\left(B_{i} \mid \Sigma_{h}\right)=\int \mathbb{P}\left(B_{i} \mid \Sigma_{i}\right) \mathbb{P}\left(\Sigma_{i} \mid \Sigma_{h}\right)\mathrm{d} \Sigma_{i}$. As we defined in equation \eqref{eq:Psigma}, $\mathbb{P}\left(\Sigma_{i} \mid \Sigma_{h}\right)$ is the density function of an inverse Wishart distribution $IW\left(\Sigma_{h}, \nu_{0}\right)$. Therefore, the most challenging part in deriving Theorem \ref{Theorem1} is to measure the density of a sparse BN $B_{i}$ given a covariance matrix $\Sigma_{i}$, \emph{i.e.}, $\mathbb{P}\left(B_{i} \mid \Sigma_{i}\right)$. Fortunately, \cite{dawid1993hyper} introduces the hyper inverse Wishart (HIW) distribution to address this issue in the context of decomposable undirected graphs. Given a decomposable undirected graph $\widetilde{G}_i$, the role of HIW distribution is to limit the parameter space of an inverse Wishart distribution according to this graph. In our case, $\mathbb{P}\left(\widetilde{G}_{i} \mid \Sigma_{i}\right) \mathbb{P}\left(\Sigma_{i} \mid \Sigma_{h}\right)$ is an HIW distribution, denoted as $HIW(\Sigma_{i}, \Sigma_{h},\widetilde{G}_{i})$. Marginalizing out $\Sigma_{i}$, we obtain the normalizing constant
\[
h(\Sigma_{h},\widetilde{G}_{i})=\int \mathbb{P}\left(\widetilde{G}_{i} \mid \Sigma_{i}\right) \mathbb{P}\left(\Sigma_{i} \mid \Sigma_{h}\right) \mathrm{d} \Sigma_{i}
\]
and have the identity $\mathbb{P}\left(B_{i} \mid \Sigma_{h}\right)=\mathbb{P}\left(\widetilde{G}_{i} \mid \Sigma_{h}\right)=h(\Sigma_{h},\widetilde{G}_{i})$.
\end{remark}

\begin{theorem}
\label{Theorem2}
For the target structure $B_i$ corresponding to dataset $D_i$,
\begin{equation}
\mathbb{P}\left(D_{i} \mid B_{i}, \Sigma_{h}\right)=\prod_{k=1}^{p} \frac{\mathbb{P}\left(D_{i}^{\left\{X_{k}\right\} \cup \Pi_{k}} \mid \Sigma_{h}\right)}{\mathbb{P}\left(D_{i}^{\Pi_{k}} \mid \Sigma_{h}\right)},
\label{eq:Pdbsigma}
\end{equation}
where $\Pi_{k}$ are the parents of $X_{k}$, $D_{i}^{\left\{X_{k}\right\} \cup \Pi_{k}}$ is the dataset $D_{i}$ restricted to $X_{k}$ and $\Pi_{k}$, and $D_{i}^{\Pi_{k}}$ is the dataset $D_{i}$ restricted to $\Pi_{k}$. Each term in equation \eqref{eq:Pdbsigma} is of the form
\begin{equation}
\mathbb{P}\left(D_{i}^{C} \mid \Sigma_{h}\right)=\frac{\left|\Sigma_{hCC}\right|^{\nu_{0} / 2} \Gamma_{p}\left(\frac{\nu_{0}+n_{i}}{2}\right)}{\pi^{n_{i} p / 2}\left|\Sigma_{hCC}+n_{i} S_{iCC}\right|^{\left(\nu_{0}+n_{i}\right) / 2} \Gamma_{p}\left(\frac{\nu_{0}}{2}\right)},
\label{eq:Pdsigmah}
\end{equation}
where $\Sigma_{hCC}$ is the sub-matrix of $\Sigma_{h}$, and $S_{iCC}$ is the sub-matrix of the sample covariance matrix $S_{i}$.
\end{theorem}

Finally, multiplying $\mathbb{P}\left(D_{i} \mid B_{i}, \Sigma^{h}\right)$, $\mathbb{P}\left(B_{i} \mid \Sigma_{h}\right)$, and $\mathbb{P}\left(\Sigma_{h}\right)$, we obtain the complete likelihood function $\mathbb{P}\left(\Sigma_{h}, \mathcal{D}, \mathcal{B}\right)$ in \eqref{eq:completell}. 

\subsection{The conditional distribution}
\label{subsec:conditional_distr}

The EM algorithm requires computing the expectation of the complete log-likelihood function with respect to the conditional distribution $\mathbb{P}\left(\Sigma_{h} \mid \mathcal{B}, \mathcal{D}\right)$. As a first step, Theorem \ref{Theorem3} gives its density function in a closed form.

\begin{theorem}
\label{Theorem3}
Suppose that $\tilde{G}_i$ is the decomposable undirected graph generated by $B_i$, and then we have
\[
\mathbb{P}\left(\Sigma_{h} \mid \mathcal{B}, \mathcal{D}\right)=\prod_{i=1}^{m} \frac{\left|\Sigma_{h}\right|^{\nu_{0} / 2} \cdot h\left(\Sigma_{h}+n_{i} S_{i}, \widetilde{G}_{i}\right)}{\left|\Sigma_{h}+n_{i} S_{i}\right|\left(\nu_{0}+n_{i}\right) / 2},
\]
where the term $h(\cdot,\widetilde{G}_{i})$ is given in equation \eqref{eq:hfunc}.
\end{theorem}

\subsection{The EM algorithm}
\label{subsec:em_algorithm}

To maximize $\mathcal{L}\left(\mathcal{B},\mathcal{D}\right)$ with respect to $\mathcal{B}$, the EM algorithm starts with an initial value $\mathcal{B}^{(0)}$, and then repeats the following two steps until convergence:
\begin{itemize}
\item \textbf{E-step}: Define the function
	\begin{equation}
	\mathcal{Q}\left(\mathcal{B} \mid \mathcal{B}^{(t-1)}, \mathcal{D}\right)=\int \log \mathbb{P}\left(\Sigma_h,\mathcal{B},\mathcal{D}\right) \mathbb{P}\left(\Sigma_{h} \mid \mathcal{B}^{(t-1)}, \mathcal{D}\right) \mathrm{d} \Sigma_{h}.\label{eq:Qfunc}
	\end{equation}
\item \textbf{M-step}: Solve the optimization problem below to generate the next iterate $\mathcal{B}^{(t)}$:
	\[
	\mathcal{B}^{(t)}=\underset{\mathcal{B}}{\arg\max}\ \mathcal{Q}\left(\mathcal{B} \mid \mathcal{B}^{(t-1)}, \mathcal{D}\right).
	\]
\end{itemize}

The basic convergence properties of the EM algorithm were established by \cite{boyles1983convergence} and \cite{wu1983convergence}, and \cite{friedman2013bayesian} has proved that the algorithm converges when there is no further improvement in the objective score. However, in our setting, the $\mathcal{Q}$-function in \eqref{eq:Qfunc} is still intractable due to the high-dimensional integration. Instead, we view the integral in \eqref{eq:Qfunc} as an expectation, and then use Monte Carlo samples to approximate it. This leads to the MCEM algorithm, which is a practical computing method for the EM algorithm.

It should be pointed out that the convergence properties of MCEM is more complicated than ordinary EM. Some efforts have been made to establish the convergence properties of MCEM; see for example \cite{chan1995monte} and \cite{fort2003convergence}. We describe our implementation of the MCEM algorithm for MTGBN in Section \ref{sec:computing}.

\section{Computing method}
\label{sec:computing}

\subsection{MC E-step}

The basic idea of MCEM is to replace the exact $\mathcal{Q}\left(\mathcal{B} \mid \mathcal{B}^{(t-1)}, \mathcal{D}\right)$ function in \eqref{eq:Qfunc} by the following unbiased estimator:
\begin{equation}
\widetilde{\mathcal{Q}}\left(\mathcal{B} \mid \mathcal{B}^{(t-1)}, \mathcal{D}\right) = \frac{1}{N} \sum_{l=1}^{N} \log \mathbb{P}\left(\Sigma_{h}^{l}, \mathcal{B}, \mathcal{D}\right),
\label{eq:Qtilde}
\end{equation}
where $\{\Sigma_{h}^{l}\}_{l=1}^{N}$ is a Monte Carlo sample of $\mathbb{P}\left(\Sigma_{h} \mid \mathcal{B}^{(t-1)}, \mathcal{D}\right)$.
Therefore, the key to the MC E-step is a proper sampling algorithm for $\mathbb{P}\left(\Sigma_{h} \mid \mathcal{B}^{(t-1)}, \mathcal{D}\right)$.

There are two main challenges that need to be addressed: (1) the positive-definiteness of $\Sigma_{h}$ must be maintained; (2) the sampling scheme needs to be fast and efficient. For the first challenge, we use the Cholesky decomposition to transform $\Sigma_{h}$ into an unrestricted vector $V$, and implement the sampling on the unrestricted space. For the second problem, we employ the HMC sampler \citep{neal2011mcmc} for efficient sampling, which is a powerful tool for Bayesian computation, especially in high-dimensional and other complicated scenarios. It has been empirically justified that HMC can offer greater computational efficiency than the traditional Gibbs sampling and Metropolis--Hastings algorithm.

Specifically, we factorize $\Sigma_{h}$ using the Cholesky decomposition $\Sigma_{h}=LL^{T}$, where $L$ is a lower-triangular matrix with positive diagonal elements (\emph{i.e.}, $l_{ii}>0$ for $i=1,2,\ldots,p$ and $l_{ij}=0$ for $i<j$). To further remove the constraints on $l_{ii}$, we work on the reparameterized variables\footnote{The similar method is also used by the Stan software: \url{https://mc-stan.org/docs/2_28/reference-manual/covariance-matrices-1.html}.}, $\log(l_{ii})$, and let $V=(v_{ij})$ be a lower-triangular matrix such that $v_{ii}=\log(l_{ii})$ for $i=1,2,\ldots,p$ and $v_{ij}=l_{ij}$ for $i>j$. In this way, $V$ collects all unrestricted variables for $\Sigma_{h}$. HMC requires evaluating both $\log \mathbb{P}(V|\mathcal{B}^{(t-1)},\mathcal{D})$ and $\nabla_{V} \log \mathbb{P}(V|\mathcal{B}^{(t-1)},\mathcal{D})$, and we summarize the results in Theorem \ref{thm:distr_v}.

\begin{theorem}
\label{thm:distr_v}
Let $\mathbf{diag}\{A\}$ be the diagonal of a matrix $A$, and $\mathbf{lower}\{A\}$ be the lower-triangular part of $A$ excluding the diagonal. Denote by $I$ the identity matrix, and $I_{C}$ a matrix constructed by selecting rows of $I$ indexed by $C$. Let $u\circ v$ be the elementwise multiplication of two vectors $u$ and $v$. Then the density function of $V$ given $\mathcal{B}^{(t-1)}$ and $\mathcal{D}$ is
\begin{equation}
	\mathbb{P}(V|\mathcal{B}^{(t-1)},\mathcal{D})=\prod_{i=1}^{m} \frac{\left|\Sigma_{h}\right|^{\nu_{0} / 2} \cdot h\left(\Sigma_{h}+n_{i} S_{i}, \mathcal{B}_{i}^{(t-1)}\right)}{\left|\Sigma_{h}+n_{i} S_{i}\right|\left(\nu_{0}+n_{i}\right) / 2}\cdot2^{p}\prod_{i=1}^{p}l_{ii}^{p-i+2},\label{eq:density_v}
\end{equation}
and the gradient of the log-density function is
\begin{equation}
\begin{aligned}
	\mathbf{diag}\{\nabla_{V} \log \mathbb{P}(V|\mathcal{B}^{(t-1)},\mathcal{D})\}&=\mathbf{diag}\{\widetilde{L}+q(m,\nu_{0},p)\cdot I\} \circ \mathbf{diag}\{L\},\\
	\mathbf{lower}\{\nabla_{V} \log \mathbb{P}(V|\mathcal{B}^{(t-1)},\mathcal{D})\}&=\mathbf{lower}\{\widetilde{L}\},
\end{aligned}
\label{eq:deriv_v}
\end{equation}
where
\begin{align*}
	\widetilde{L}&=g(\nu_{0}+n_{i},\Sigma_{h}+n_{i} S_{i}, \mathcal{B}^{(t-1)})-f(\nu_{0}+n_{i},\Sigma_{h}+n_{i} S_{i}),\\
	g(\nu,\Sigma,\mathcal{B})&=2\nu\left\{\sum_{C \in \mathcal{C}} I_{C}^{\mathrm{T}}(\Sigma_{CC})^{-1}I_{C}L-\sum_{S \in \mathcal{S}}I_{S}^{\mathrm{T}}(\Sigma_{SS})^{-1}I_{S}L\right\},\\
	f(\nu,\Sigma)&=\nu \cdot \Sigma^{-1}L,\\
	q(m,\nu,p)&=\frac{p(p+2m\nu+3)}{2}.
\end{align*}
\end{theorem}
Based on equations \eqref{eq:density_v} and \eqref{eq:deriv_v}, we can use the HMC algorithm to obtain a set of sample $V$ through iterations, denoted as $\{V^{(l)}\}_{l=1}^{N}$. Then by transfroming $V$ back to $\Sigma_{h}$, we can obtain $\{\Sigma_{h}^{(l)}\}_{l=1}^{N}$. Details of the derivation and the algorithm are given in Appendix \ref{Appendix d}.

\subsection{MC M-step}

Finally in the MC M-step, we maximize the approximate $\mathcal{Q}$ function $\widetilde{\mathcal{Q}}\left(\mathcal{B} \mid \mathcal{B}^{(t-1)}, \mathcal{D}\right)$ as in \eqref{eq:Qtilde}, and take $\mathcal{B}^{(t)}$ to be the maximizer. We use the max-min hill-climbing algorithm (MMHC, \citealp{tsamardinos2006max}) to search the BN structures, which combines ideas from local learning, constraint-based methods, and search-and-score techniques. It has been empirically justified that on average MMHC outperforms several prototypical algorithms \citep{wang2018analysis}, such as PC \citep{kalisch2007estimating}, sparse candidate \citep{friedman2013learning}, greedy equivalence search \citep{ramsey2017million}, and greedy search \citep{chickering2002optimal}.

It is observed that $\{B^{(t-1)}_{i}\}_{i=1}^{m}$ are independent with each other in the objective function $\widetilde{\mathcal{Q}}\left(\mathcal{B} \mid \mathcal{B}^{(t-1)}, \mathcal{D}\right)$. Therefore, we can apply MMHC $m$ times on the $m$ independent sub-problems:
\[
B_{i}^{(t)}=\underset{B_{i}}{\arg\max}\ \sum_{l=1}^{N}\left\{\log \mathbb{P}(D_{i} \mid B_{i},\Sigma_{h}^{l})+\log \mathbb{P}(B_{i} \mid \Sigma_{h}^{l})\right\}.
\]
The overall MCEM algorithm of MTGBN is thereby given in Algorithm \ref{mcem}.
\begin{algorithm}[h]
	\caption{MCEM algorithm of MTGBN}
	\begin{algorithmic}[1]\label{mcem}
		\REQUIRE $\mathcal{D}$: Observational datasets; $\mathcal{B}^{(0)}$: Initial BN structures; $\varepsilon$: Tolerance.
		\ENSURE $\hat{\mathcal{B}}$: Estimated BN structures.
		\FOR{$t=1, \ldots, T$}

			\STATE \textbf{MC E-step:}
			\STATE \quad $\{\Sigma_{h}^{(l)}\}_{l=1}^{N}$ $\leftarrow$ \textbf{HMC} ($\mathcal{D}$, $\mathcal{B}^{(t-1)}$), see Appendix \ref{Appendix d} for details.
			\STATE \quad  Define function $\widetilde{\mathcal{Q}}\left(\mathcal{B} \mid \mathcal{B}^{(t-1)}, \mathcal{D}\right)$ by equation \eqref{eq:Qtilde}.\smallskip{}

			\STATE \textbf{MC M-step:}
			\STATE \quad\textbf{for} $i=1, \ldots, m$ \textbf{do}
			\STATE \qquad $B_{i}^{(t)}$ $\leftarrow$ \textbf{MMHC} ($D_{i}, B_{i}^{(t-1)}$, $\{\Sigma_{h}^{(l)}\}_{l=1}^{N}$).
			\STATE \qquad MMHC is a heuristic algorithm to detect BN structures \citep{tsamardinos2006max}.
			\STATE \quad\textbf{end for}
			\STATE \quad Set $\mathcal{B}^{(t)}=\{B_{i}^{(t)}\}_{i=1}^{m}$.\smallskip{}

            \STATE \textbf{break if} $$\left|\widetilde{\mathcal{Q}}\left(\mathcal{B}^{(t)} \mid \mathcal{B}^{(t-1)}, \mathcal{D}\right)-\widetilde{\mathcal{Q}}\left(\mathcal{B}^{(t-1)} \mid \mathcal{B}^{(t-1)}, \mathcal{D}\right)\right|\leq \varepsilon.$$
		\ENDFOR
		\RETURN $\hat{\mathcal{B}}=\mathcal{B}^{(t)}$.
	\end{algorithmic}
\end{algorithm}

\section{Simulations}
\label{sec:simulation}

In this section, we empirically evaluate the finite sample performance of the proposed MTGBN method on synthetic data and benchmark network data and compare it with several competing methods.  

\subsection{Simulation design of synthetic data}\label{subsec:synthetic data}

We simulate synthetic datasets with our data generation mechanism (see Figure \ref{framework}). First, we random generate $\Sigma_{h}$ from the uniform distribution over all $p \times p$ positive definite matrices with unit trace, using an approach similar to \cite{mittelbach2012sampling}. Then given $\Sigma_{h}$, we sample $m$ intermediate matrices $\{\tilde{\Sigma}_{l}\}_{l=1}^{m}$ from $I W\left(\Sigma_{h}, \nu_{0}\right)$, so each $\tilde{\Sigma}_{l}$ is a $p \times p$ positive definite matrix.

The $m$ undirected graph structures  $\left\{G_{l}\right\}_{l=1}^{m}$ are defined as follows. For each $l=1,\dots,m$, let $\tilde{\Omega}_l=(\tilde{\omega}_{ij,l})=\tilde{\Sigma}_l^{-1}\in\mathbb{R}^{p\times p}$, $V=\{1, \ldots, p\}$, and define $E_l=\{(i, j): i \neq j, |\tilde{\omega}_{ij,l}| > t_l\}$, where $t_l>0$ is a threshold to control the density of $E_l$. Let $\Omega_l=(\omega_{ij,l})$ be a matrix such that  $\omega_{ii,l}=\tilde{\omega}_{ii,l}$, $\omega_{ij,l}=\tilde{\omega}_{ij,l}$ if $(i,j)\in E_l$, and $\omega_{ij,l}=0$ otherwise. We assume that $\Omega_l$ is still positive definite, which can be achieved by repeatedly sampling $\tilde{\Sigma}_l$ until the condition is met. Then the graphs are given by $G_l=(V,E_l)$.

Denote by $|E_l|$ the cardinality of $E_l$, so $d_l=|E_l|/p^2$ represents the density of $G_l$, and is dependent on the threshold $t_l$. By adjusting $t_l$, we can obtain $G_l$ with different density levels. In fact, the zero elements in $\Omega_l$ imply the conditional independence structure of the variables in $G_l$ \citep{dawid1993hyper}.

Finally, we generate $m$ independent datasets $\left\{D_{l}\right\}_{l=1}^{m}$ by sampling each $D_{l}=\left\{\bx_{l 1}, \ldots, \bx_{l n_{l}}\right\}$ from a multivariate normal distribution $\mathcal{N}\left(0, \Sigma_{l}\right)$, where $\Sigma_l=\Omega_l^{-1}$, and $n_l$ is the sample size for the $l$-th dataset.

To evaluate and compare the performance of the MTGBN method, we also learn each network independently with single-task learning (SIG for short), and learn a common network structure for all contexts (AVG for short), \emph{i.e.}, an ``average'' network that treats all subjects as samples in one task. AVG was previously used as a baseline in \cite{oyen2012leveraging}. To compare the performance of MTGBN, SIG, and AVG, especially on the differences between the structures, we use edge error, edge precision, edge F-score, and edge distance as the primary evaluation measures.

Let $\left\{B_{output, l}\right\}_{l=1}^{m}$ represent the $m$ BN structures obtained from MTGBN, SIG, or AVG method, and $\left\{G_{output, l}\right\}_{l=1}^{m}$ be the undirected graph structures corresponding to $\left\{B_{output, l}\right\}_{l=1}^{m}$. In the experiment of synthetic data, since the generated networks $\left\{G_{l}\right\}_{l=1}^{m}$ are undirected graphs, we mainly evaluate the network structures without considering the edge orientation, which we call the adjacency measure. To compute the evaluation metrics, we first calculate four basic statistics: true positives (TP), the number of edges that are both in $G_{output, 1}$ and $G_{1}$; false positives (FP), the number of edges that are present in $G_{output, 1}$ but not in $G_{ 1}$; false negatives (FN), the number of edges that are present in $G_{ 1}$ but not in $G_{output, 1}$; and true negatives (TN), the number of vertex pairs that are neither edges in $G_{ 1}$ nor in $G_{output, 1}$. Then the edge adjacency error, edge adjacency precision, edge adjacency F-score, and edge adjacency distance are defined as: 

\begin{itemize}
    \item \textbf{Error:}
     \begin{equation}
     \text {\emph{Error}}=\frac{FP+F N}{T P+T N+F P+F N}. \label{eq:error}
    \end{equation}
    \item \textbf{Precision:}
    \[
     \text { \emph{Precision} }=\frac{T P}{T P+F P}.
    \]
     \item \textbf{F-score:}
     \[
      \text { \emph{F-score}}=2 \cdot \frac{\text { \emph{Precision} } \cdot \text {\emph{Recall} }}{\text { \emph{Precision} }+\text {\emph{Recall}}},
     \]
     where $\text { \emph{Recall}}=TP/(TP+FN)$.
     \item \textbf{Edge distance:} Given two structures, this measure is the minimum number of edges needed to convert the learned graph into the true one.
\end{itemize}

We compare the performance of MTGBN, SIG, and AVG on the following three aspects.

\textbf{Performance v.s. sample size:} We fix the task number $m=10$, number of nodes $p=15$, density level $d_l=0.3$, and compare the model performance of network structure learning for different sample size $n_l=50, 100, 150, 200, 250, 300$.

\textbf{Performance v.s. task number:} We fix the number of nodes $p=15$, density level $d_l=0.3$, sample size $n_l=250$, and consider different task number $m=2, 4, 6, 8, 10$.

\textbf{Performance v.s. density level:} We fix the task number $m=10$, number of nodes $p=15$, and consider a combination of different sample size $n_l=100,200,300$ and density levels ranging from $0.2$ to $0.6$.

The evaluation metrics for one simulation run are averaged over the $m$ datasets, and we repeat each experiment $10$ times. The final evaluation metrics are thereby averaged over the $10$ simulation runs.

\subsection{Simulation design of benchmark network data}

In this experiment, we select benchmark networks from the Bayesian Network Repository \citep{constantinou2020bayesys}. We choose three medium-sized benchmark networks and generate low sample size data to validate our assumption that the proposed MTGBN method can improve the understanding of the underlying structure and generalize better even at small sample sizes. The information of the selected benchmark networks is provided in Table \ref{tab:benchmark}.
\begin{table}[h]
\centering
\caption{Facts of benchmark networks from the Bayesian Network Repository}
\begin{tabular}{@{}lll@{}}
\toprule
Benchmark Networks & Nodes & Arcs \\
\midrule
ECOLI70 & 46 & 70  \\
MAGIC-NIAB & 44 & 66 \\
MAGIC-IRRI & 64 & 102 \\
\bottomrule
\end{tabular}
\label{tab:benchmark}
\end{table}

To obtain a set of related networks, we apply perturbations to the given benchmark network to create similar but different networks. Denote by $B_{truth}$ the benchmark BN with an adjacency matrix $A_{t}=(a^{t}_{ij})\in\mathbb{R}^{p\times p}$, where $a^{t}_{ij}=1$ if there is an edge from node $i$ to node $j$, and $a^{t}_{ij}=0$ otherwise, $i,j=1,\ldots,p$. Then the generated BN, denoted by $B_{gen}$, is constructed in the following way. Let $A_{g}$ be the adjacency matrix of $B_{gen}$, and define the perturbation level as the ratio of the number of changed edges in $A_{g}$ to the number of all possible edges. For example, if the perturbation level is $30\%$, we first set $A_{g}=A_{t}$, and then randomly select $30\%$ of the off-diagonal elements $a^{t}_{ij}(i\neq j)$ in $A_{t}$. For each selected element $a^{t}_{ij}$, if $a^{t}_{ij}=0$, then we set $a^{g}_{ij}=1$ (adding edge); if $a^{t}_{ij}=1$, then with probability $1/2$ we set $a^{g}_{ij}=0$ (deleting edge), otherwise let $a^{g}_{ji}=1$ (reversing edge). For each benchmark BN, we generate $m$ networks by such a perturbation scheme, and in our experiment we set $m=10$. Each set of generated networks can be viewed as a class of tasks sharing the same prior, since they are generated from the same benchmark BN. Then we simulate data for each generated BN with a sample size $n_l=250, l=1,\dots,m$.

We mainly compare the edge arrowhead error under different network perturbation levels in the benchmark network experiment. The edge arrowhead error is the same as equation \eqref{eq:error}, except that TP, FP, FN, and TN are calculated considering the edge orientation. Same as Section \ref{subsec:synthetic data}, we repeat the experiment 10 times and compute the averaged errors for MTGBN, SIG, and AVG.

\subsection{Simulation results} 

\begin{figure}
    \centering
	  \subfloat[]{
        \includegraphics[width=0.37\linewidth]{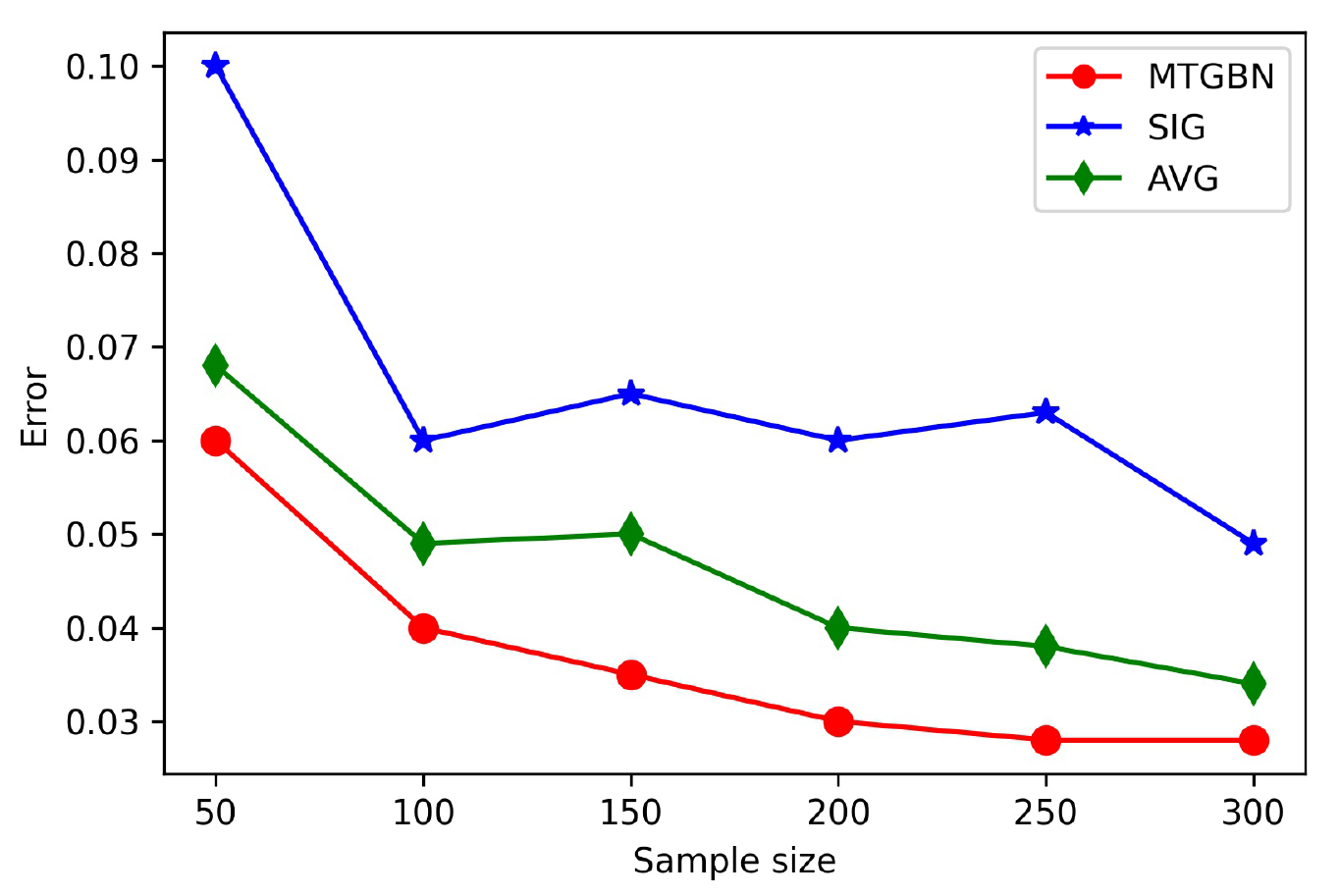}}
    \label{2a}
	  \subfloat[]{
        \includegraphics[width=0.37\linewidth]{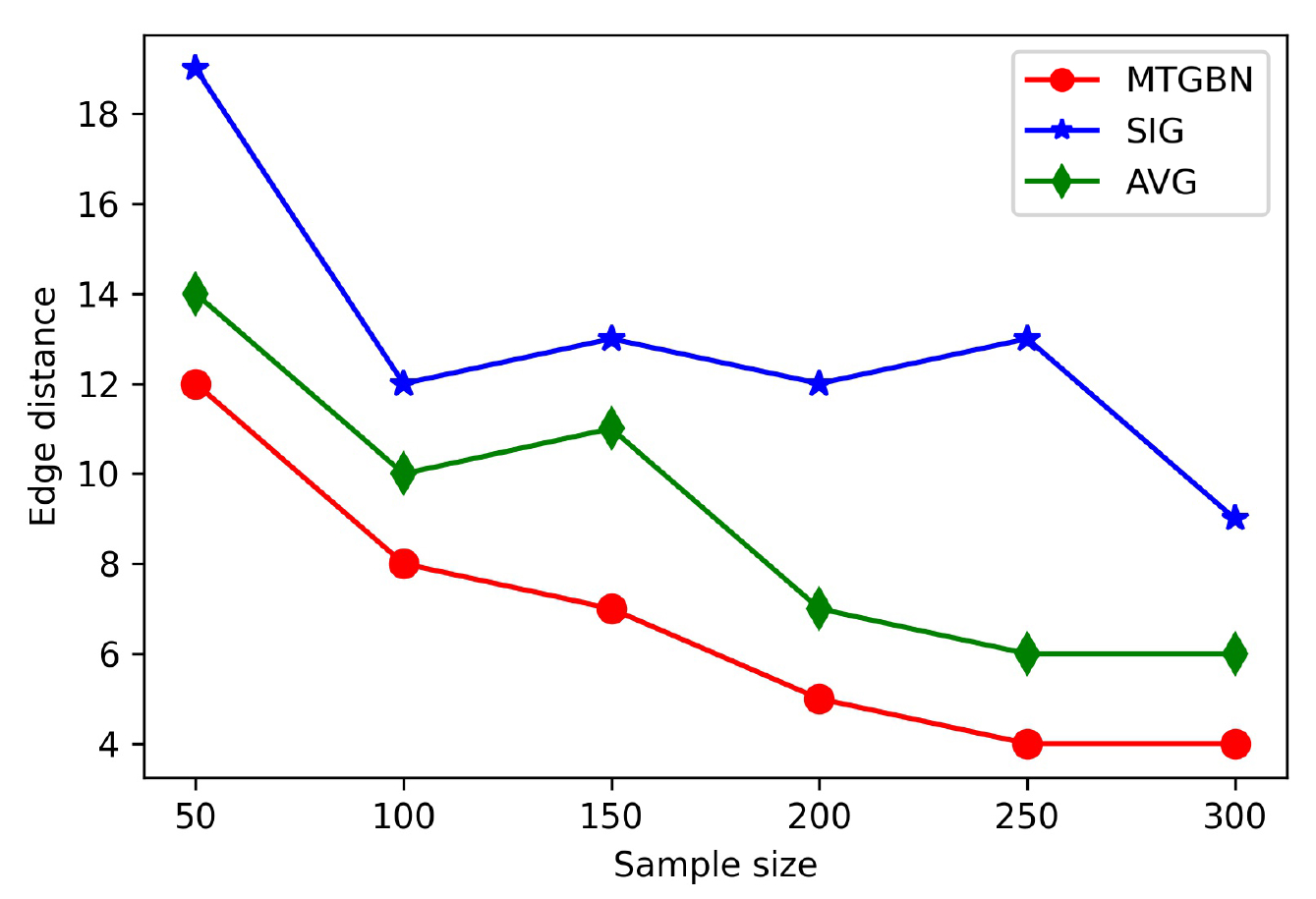}}
    \label{2b}\\
	  \subfloat[]{
        \includegraphics[width=0.37\linewidth]{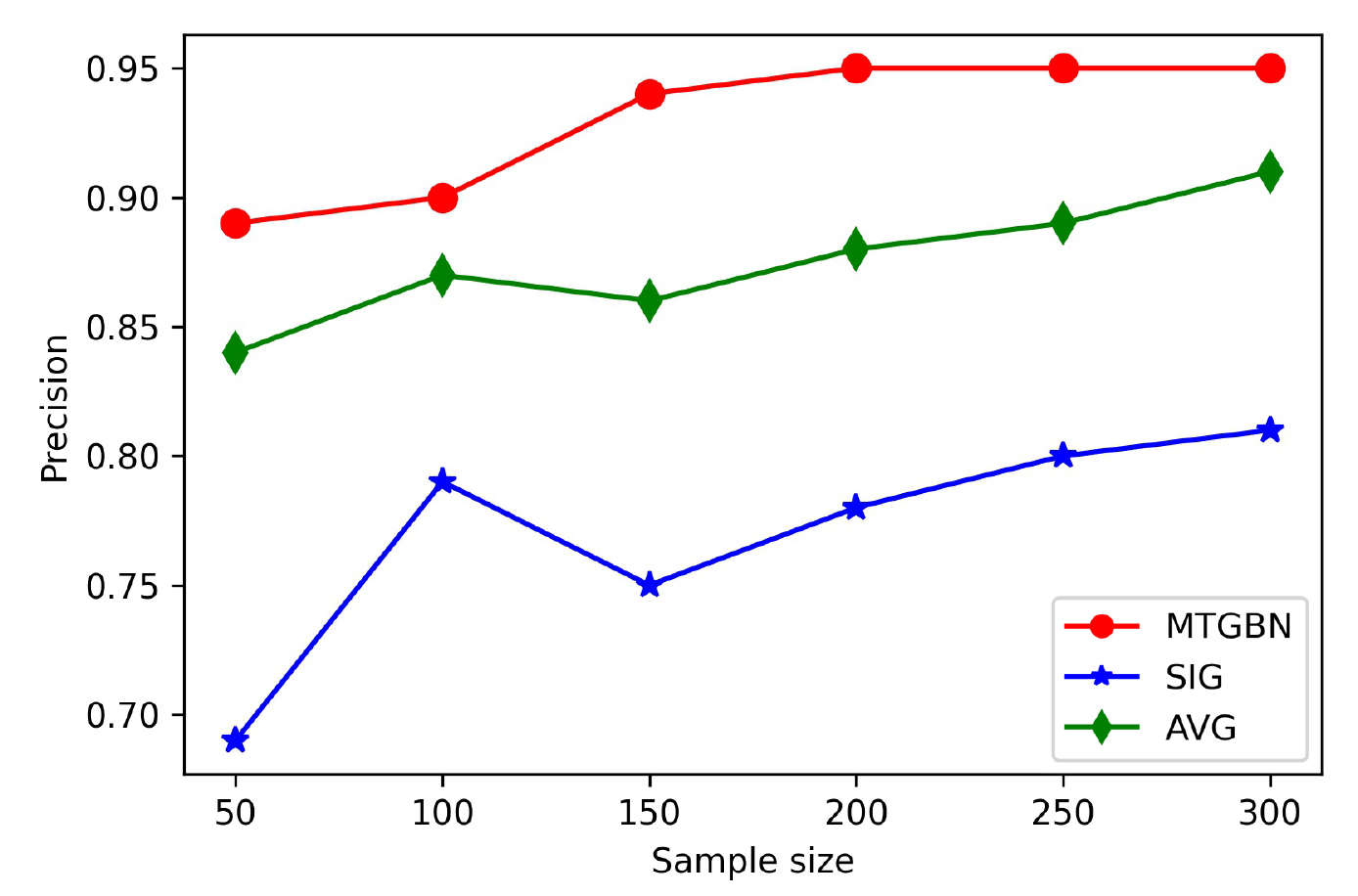}}
    \label{2c}
	  \subfloat[]{
        \includegraphics[width=0.37\linewidth]{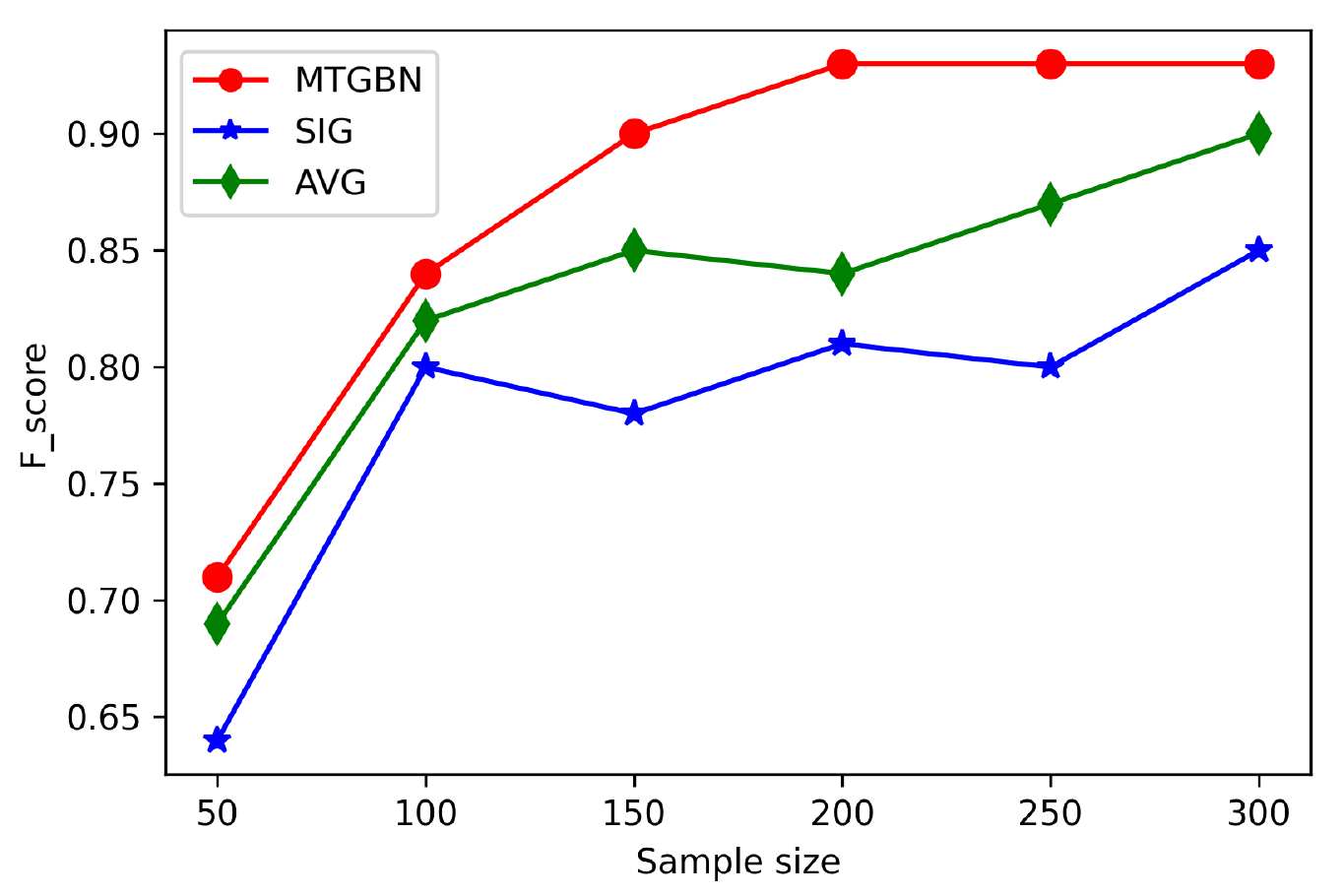}}
     \label{2d}
	  \caption{Simulation results for synthetic data under different sample sizes. (a) Average edge adjacency error (lower is better); (b) Average edge adjacency distance (lower is better); (c) Average edge adjacency precision (higher is better); (d) Average edge adjacency F-score (higher is better).}
	  \label{fig:preformance_datasize}
\end{figure}

In the synthetic data experiment, we first evaluate the performance of the proposed MTGBN method under different sample sizes, with the results shown in Figure \ref{fig:preformance_datasize}. It is clear that MTGBN significantly outperforms the other two methods. The average edge adjacency error of MTGBN under different sample sizes is in the range of 0.03~-~0.06, whereas for SIG, even using five times the number of training data (sample size $=250$), its average edge adjacency error is still larger than MTGBN (sample size $=50$). The same trend is observed in the edge adjacency distance. For the other two metrics, MTGBN consistently results in higher average edge adjacency precision and F-score than AVG and SIG. Compared with SIG, the average edge adjacency precision of MTGBN is also more stable under different sample sizes, and the best result is achieved at the sample size of 200. Such findings show that the proposed MTGBN method can produce satisfactory results using a small amount of data, thereby demonstrating its usefulness in cases where data are expensive or hard to collect.

\begin{figure}
    \centering
	  \subfloat[]{
        \includegraphics[width=0.37\linewidth]{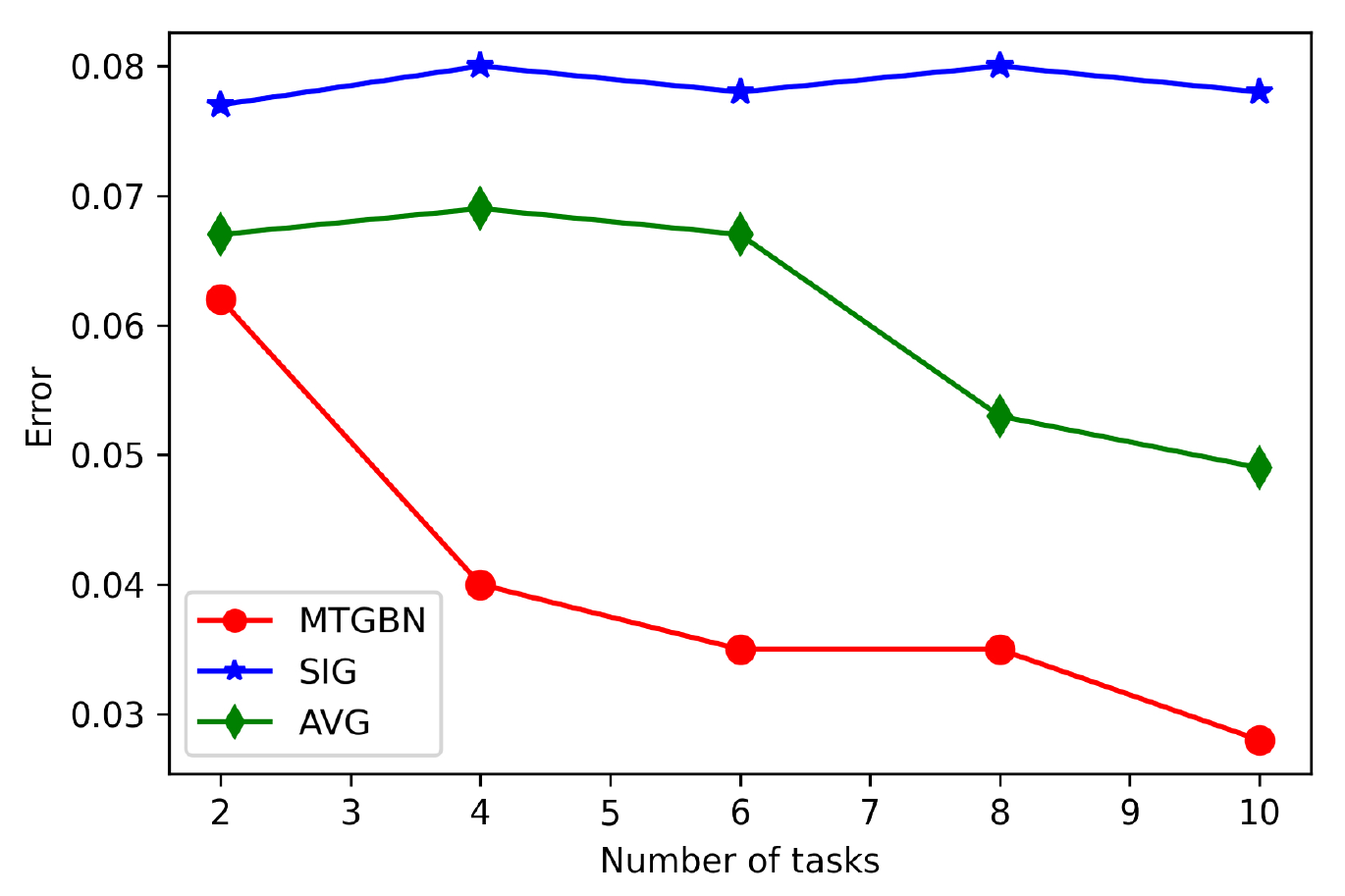}}
    \label{3a}
	  \subfloat[]{
        \includegraphics[width=0.37\linewidth]{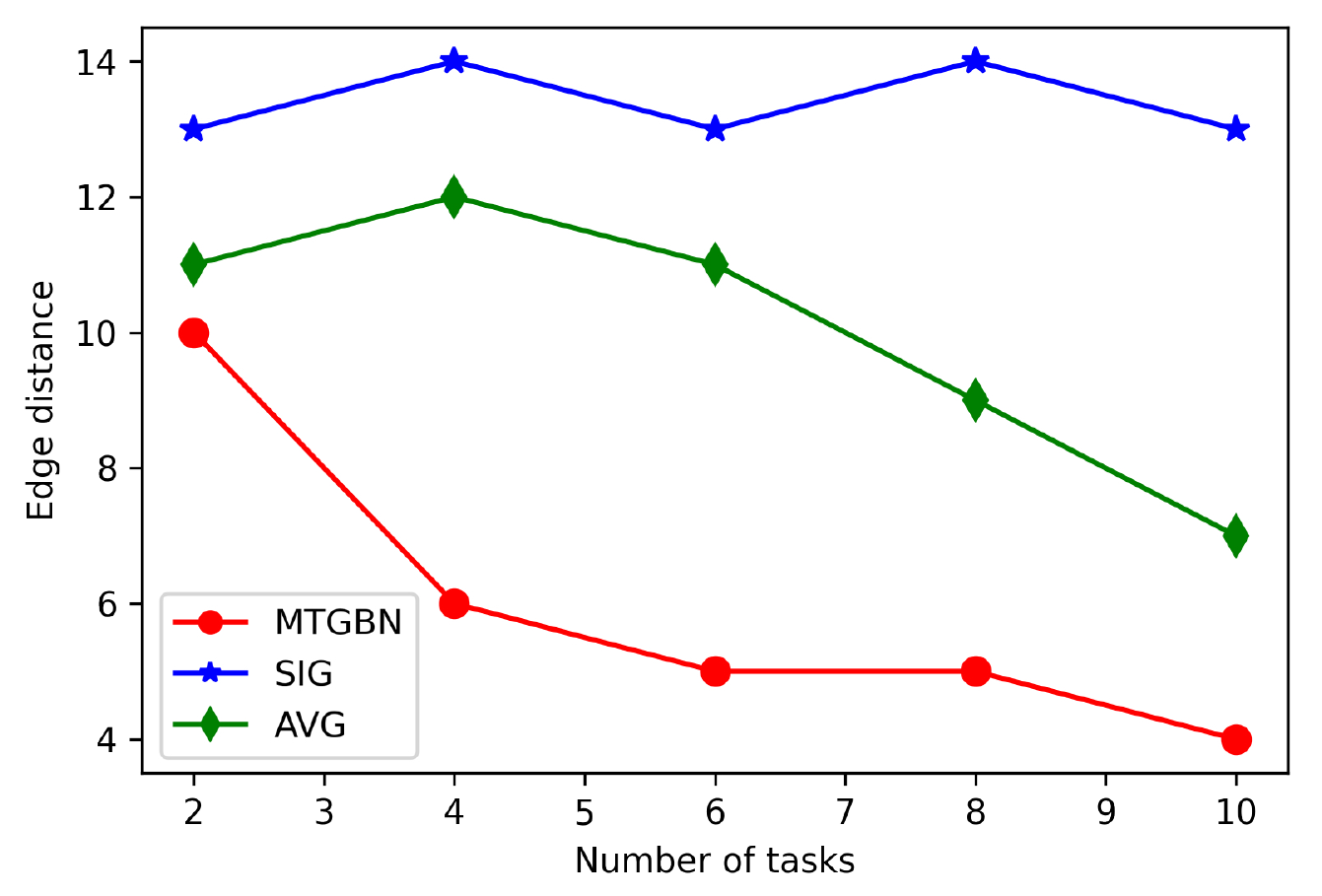}}
    \label{3b}\\
	  \subfloat[]{
        \includegraphics[width=0.37\linewidth]{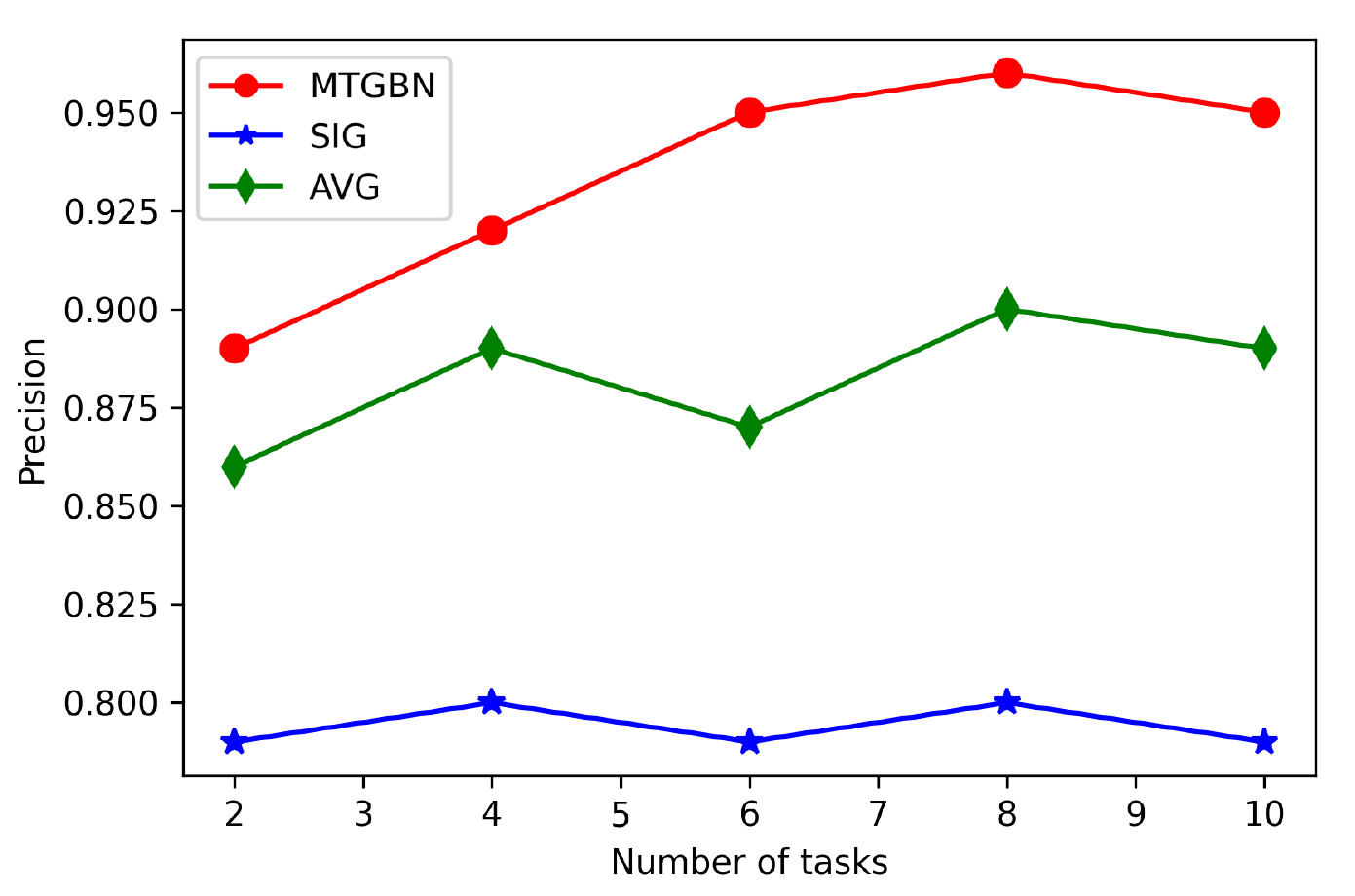}}
    \label{3c}
	  \subfloat[]{
        \includegraphics[width=0.37\linewidth]{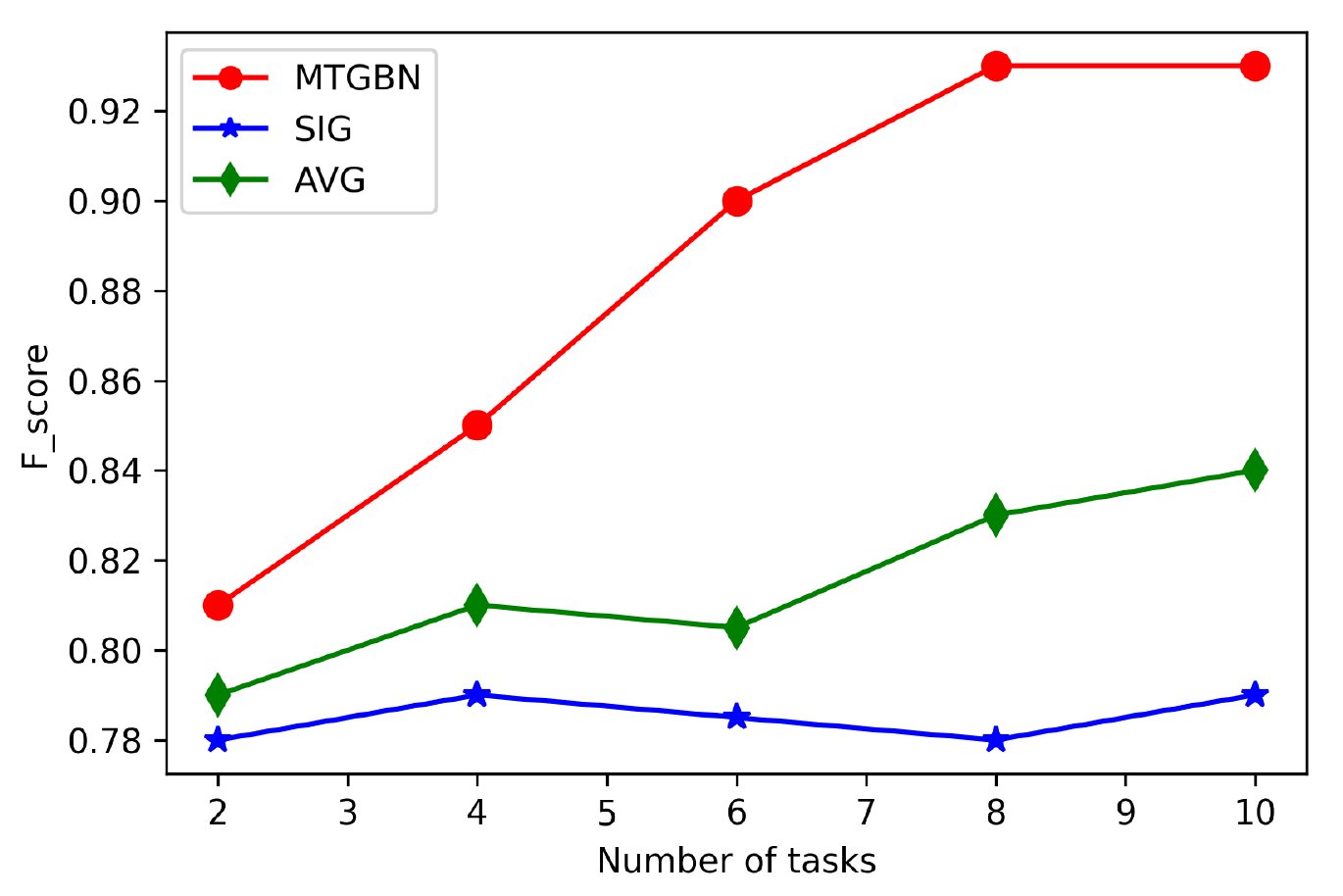}}
     \label{3d}
	  \caption{Simulation results for synthetic data under different task numbers. (a) Average edge adjacency error (lower is better); (b) Average edge adjacency distance (lower is better); (c) Average edge adjacency precision (higher is better); (d) Average edge adjacency F-score (higher is better).}
	  \label{fig:preformance_task}
\end{figure}

We also compare the performance of MTGBN, AVG, and SIG under different task numbers $m$. Figure \ref{fig:preformance_task} indicates that MTGBN performs better than AVG and SIG for all the test cases. As expected, SIG does not have better performance at a larger value of $m$, since it only utilizes the information of a single task. In contrast, MTGBN can use the information of related tasks, thus leading to increased performance. We find that MTGBN can already get a good result when $m=4$, and with the increase of $m$, the performance tends to be stable. These results are also consistent with the existing literature \citep{oyen2012leveraging}.

\begin{figure}[h]
  \centering
  {\includegraphics[width=0.9\textwidth]{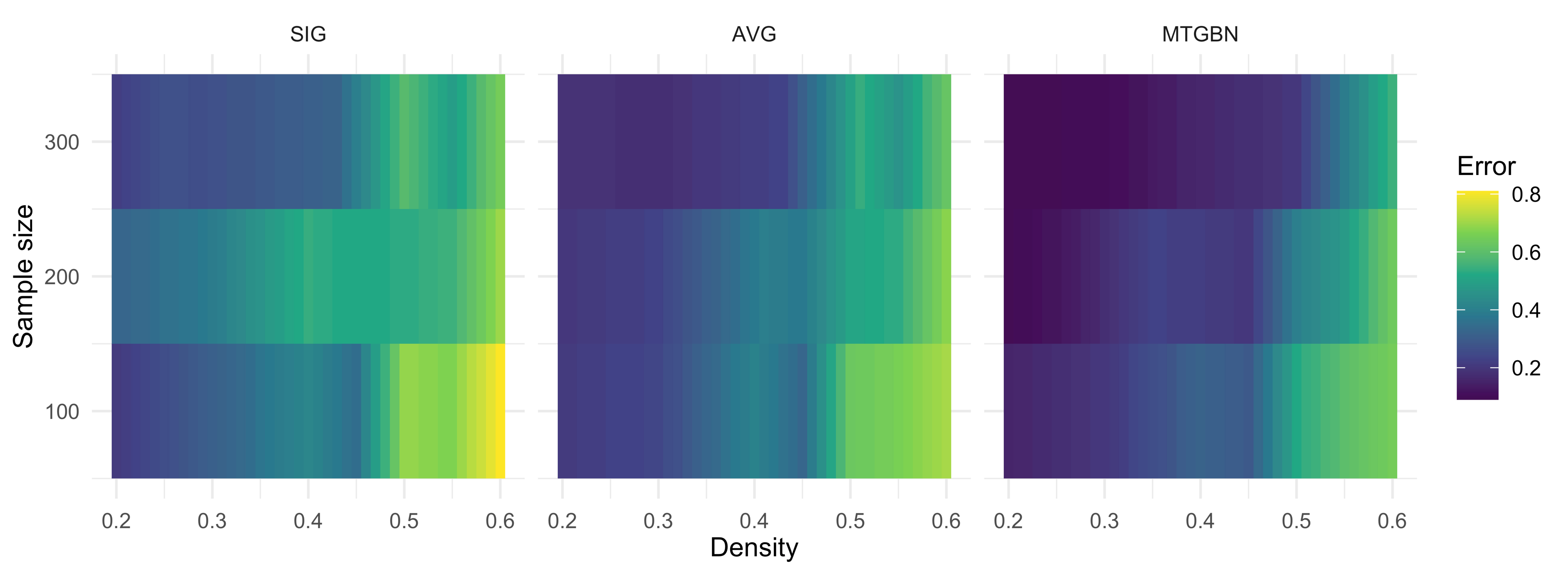}}
  \caption{Average edge adjacency errors for synthetic data for different sample sizes and density levels.}
  \label{fig:preformance_density}
\end{figure}

In addition to the sample size and task number, we find that the network density also affects the model's performance. We compare the average edge adjacency error of MTGBN, AVG, and SIG under different sample sizes and density levels, with results shown in Figure \ref{fig:preformance_density}. It can be seen that with the increase in sample size and decrease in density, the edge adjacency error of the model gets smaller. Such a phenomenon is more obvious in MTGBN. Compared with AVG and SIG, the proposed MTGBN method is especially advantageous with a small sample size ($n_l=100$) and a medium density ($d_l=0.4$).

In the experiment of the benchmark network data, we compare the average arrowhead errors for MTGBN, AVG, and SIG, and show the results in Table \ref{tab:error}. We consider four perturbation levels, $5\%$, $10\%$, $20\%$, and $30\%$, representing the proportion of possible edges that are randomly modified during simulation. Again, the comparison demonstrates that MTGBN is better than SIG and AVG all the time. Even at a high perturbation level of $30\%$, MTGBN still has decent error rates, whereas SIG and AVG do not provide good results. We also find that the more related the tasks are, the better results MTGBN can achieve.

\begin{table}[]
\centering
\caption{Average edge arrowhead errors of the learned BNs under different network perturbation levels.}
\begin{tabular}{@{}cccccc@{}}
\toprule
 &  & 5\% Perturbation & 10\% Perturbation & 20\% Perturbation & 30\% Perturbation \\ \midrule
 & SIG & 0.048 & 0.079 & 0.085 & 0.098 \\
ECOLI70 & AVG & 0.047 & 0.073 & 0.082 & 0.095 \\
 & MTGBN & \textbf{0.039} & \textbf{0.060} & \textbf{0.074} & \textbf{0.083} \\ \midrule
 & SIG & 0.053 & 0.079 & 0.093 & 0.110 \\
MAGIC-NIAB & AVG & 0.049 & 0.075 & 0.091 & 0.105 \\
 & MTGBN & \textbf{0.045} & \textbf{0.062} & \textbf{0.079} & \textbf{0.089} \\ \midrule
 & SIG & 0.091 & 0.110 & 0.124 & 0.132 \\
MAGIC-IRRI & AVG & 0.089 & 0.091 & 0.110 & 0.120 \\
 & MTGBN & \textbf{0.065} & \textbf{0.080} & \textbf{0.090} & \textbf{0.110} \\ \bottomrule
\end{tabular}
\label{tab:error}
\end{table}

\section{Applications to rs-fMRI data}
\label{sec:applications}

Recent studies have found that MDD is closely related to the changes in the brain connection network \citep{yamashita2020generalizable}, especially the changes in DMN. Several previous studies have reported abnormalities of the DMN in MDD patients and found some altered functional connections \citep{greicius2007resting,bluhm2009resting,sheline2010resting}.

Such disorder-related DMN brain functional connectivity alteration can elucidate neural mechanisms of MDD and further facilitate the diagnosis of MDD. Therefore, we can use rs-fMRI data to identify DMN brain functional connectivity patterns in MDD patients. This section will conduct such experiments and apply the proposed MTGBN method to explore connectivity alteration patterns in DMN that can facilitate MDD diagnosis.

Following the data acquisition and selection procedure introduced in Section \ref{sec:data}, we analyze the rs-fMRI data of 15 MDD patients, which are regarded as 15 related tasks. Then we can obtain a brain connection network for each patient. To study the differences in brain functional connectivity, the brain connection networks of 15 HC participants, which are viewed as another set of related tasks, are also obtained using the proposed MTGBN method. For reference, we compare MTGBN with SIG and existing standard analysis methods of rs-fMRI data: Granger causal analysis (GC, \citealp{seth2010matlab}) and dynamic causal model (DCM, \citealp{marreiros2008dynamic}).

We use MTGBN to estimate the connections between ROI nodes in three different DMN network templates: aDMN, dDMN, and vDMN. Table \ref{tab:degree} shows the difference in connection degrees of $30$ ROI nodes (including in, out, and total connection degree). Here, the connection degree represents the amount of functional connections. The greater the connection degree, the higher the effective connectivity. In aDMN, the total connection degrees are 292 for MDD and 232 for HC, indicating that MDD has 25.9\% more effective connectivity than HC. Similarly, In dDMN, MDD has 29.6\% more effective connectivity than HC with total connection degrees of 254 and 196, respectively. This increase in effective connectivity in MDD patients is also consistent with the findings of previous studies \citep{greicius2007resting,lemogne2009search,belleau2015imbalance}. The rise of DMN connectivity is an apparent robust biomarker of MDD \citep{zamoscik2014increased}, which can also be found in our results.

\begin{table}
\centering
\caption{Connection degrees of MDD patients and HC participants in three DMN network templates}
\begin{tabular}{@{}lccclccc@{}}
\toprule
Node & Total degree & In & Out & \multicolumn{1}{l}{Node} & Total degree & In & Out \\ \midrule
\textbf{aDMN (MDD)} & \multicolumn{1}{l}{} & \multicolumn{1}{l}{} & \multicolumn{1}{l}{} & \textbf{aDMN (HC)} & \multicolumn{1}{l}{} & \multicolumn{1}{l}{} & \multicolumn{1}{l}{} \\
aDMN1 & 9 & 4 & 5 & aDMN1 & 2 & 2 & 0 \\
aDMN2 & 34 & 13 & 21 & aDMN2 & 35 & 14 & 21 \\
aDMN3 & 29 & 10 & 19 & aDMN3 & 7 & 3 & 4 \\
aDMN4 & 42 & 18 & 24 & aDMN4 & 19 & 9 & 10 \\
aDMN5 & 30 & 14 & 16 & aDMN5 & 25 & 9 & 16 \\
aDMN6 & 12 & 10 & 2 & aDMN6 & 9 & 4 & 5 \\
aDMN7 & 21 & 7 & 14 & aDMN7 & 24 & 13 & 11 \\
aDMN8 & 33 & 19 & 14 & aDMN8 & 22 & 12 & 10 \\
aDMN9 & 30 & 17 & 13 & aDMN9 & 29 & 14 & 15 \\
aDMN10 & 31 & 21 & 10 & aDMN10 & 26 & 18 & 8 \\
aDMN11 & 21 & 14 & 7 & aDMN11 & 34 & 18 & 16 \\
Total & 292 & \multicolumn{1}{l}{} & \multicolumn{1}{l}{} & Total & 232 & \multicolumn{1}{l}{} & \multicolumn{1}{l}{} \\
\textbf{dDMN (MDD)} & \multicolumn{1}{l}{} & \multicolumn{1}{l}{} & \multicolumn{1}{l}{} & \textbf{dDMN (HC)} & \multicolumn{1}{l}{} & \multicolumn{1}{l}{} & \multicolumn{1}{l}{} \\
dDMN1 & 30 & 10 & 20 & dDMN1 & 22 & 2 & 20 \\
dDMN2 & 20 & 11 & 9 & dDMN2 & 12 & 10 & 2 \\
dDMN3 & 25 & 11 & 14 & dDMN3 & 16 & 5 & 11 \\
dDMN4 & 28 & 15 & 13 & dDMN4 & 27 & 17 & 10 \\
dDMN5 & 31 & 13 & 18 & dDMN5 & 18 & 12 & 6 \\
dDMN6 & 24 & 13 & 11 & dDMN6 & 23 & 13 & 10 \\
dDMN7 & 28 & 13 & 15 & dDMN7 & 15 & 9 & 6 \\
dDMN8 & 44 & 24 & 20 & dDMN8 & 36 & 17 & 19 \\
dDMN9 & 24 & 17 & 7 & dDMN9 & 27 & 13 & 14 \\
Total & 254 & \multicolumn{1}{l}{} & \multicolumn{1}{l}{} & Total & 196 & \multicolumn{1}{l}{} & \multicolumn{1}{l}{} \\
\textbf{vDMN (MDD)} & \multicolumn{1}{l}{} & \multicolumn{1}{l}{} & \multicolumn{1}{l}{} & \textbf{vDMN (HC)} & \multicolumn{1}{l}{} & \multicolumn{1}{l}{} & \multicolumn{1}{l}{} \\
vDMN1 & 7 & 2 & 5 & vDMN1 & 2 & 0 & 2 \\
vDMN2 & 29 & 11 & 18 & vDMN2 & 21 & 8 & 13 \\
vDMN3 & 23 & 14 & 9 & vDMN3 & 28 & 11 & 17 \\
vDMN4 & 26 & 13 & 13 & vDMN4 & 24 & 15 & 9 \\
vDMN5 & 18 & 8 & 10 & vDMN5 & 20 & 11 & 9 \\
vDMN6 & 26 & 10 & 16 & vDMN6 & 31 & 12 & 19 \\
vDMN7 & 30 & 16 & 14 & vDMN7 & 28 & 12 & 16 \\
vDMN8 & 19 & 9 & 10 & vDMN8 & 21 & 16 & 5 \\
vDMN9 & 38 & 21 & 17 & vDMN9 & 39 & 19 & 20 \\
vDMN10 & 22 & 15 & 7 & vDMN10 & 26 & 16 & 10 \\
Total & 238 & \multicolumn{1}{l}{} & \multicolumn{1}{l}{} & Total & 240 & \multicolumn{1}{l}{} & \multicolumn{1}{l}{} \\ \bottomrule
\end{tabular}
\label{tab:degree}
\end{table}

From Table \ref{tab:degree} we can find some ROI nodes with high connection degrees, such as aDMN4 (temporal-parietal junction), aDMN2 (posterior cingulate cortex), aDMN8 (posterior inferior parietal lobule), and aDMN10 (parahippocampal cortex) in aDMN. Specifically, aDMN4 and aDMN2 have more outgoing connections than incoming connections, whereas aDMN8 and aDMN10 have more incoming connections than outgoing connections. It indicates that aDMN4 and aDMN2 have a more essential impact on other ROI nodes, and aDMN8 is mainly affected by other ROI nodes. These findings are also consistent with the physiological discoveries of MDD \citep{li2018brain}.

\begin{figure}
  \centering
  {\includegraphics[width=0.9\textwidth]{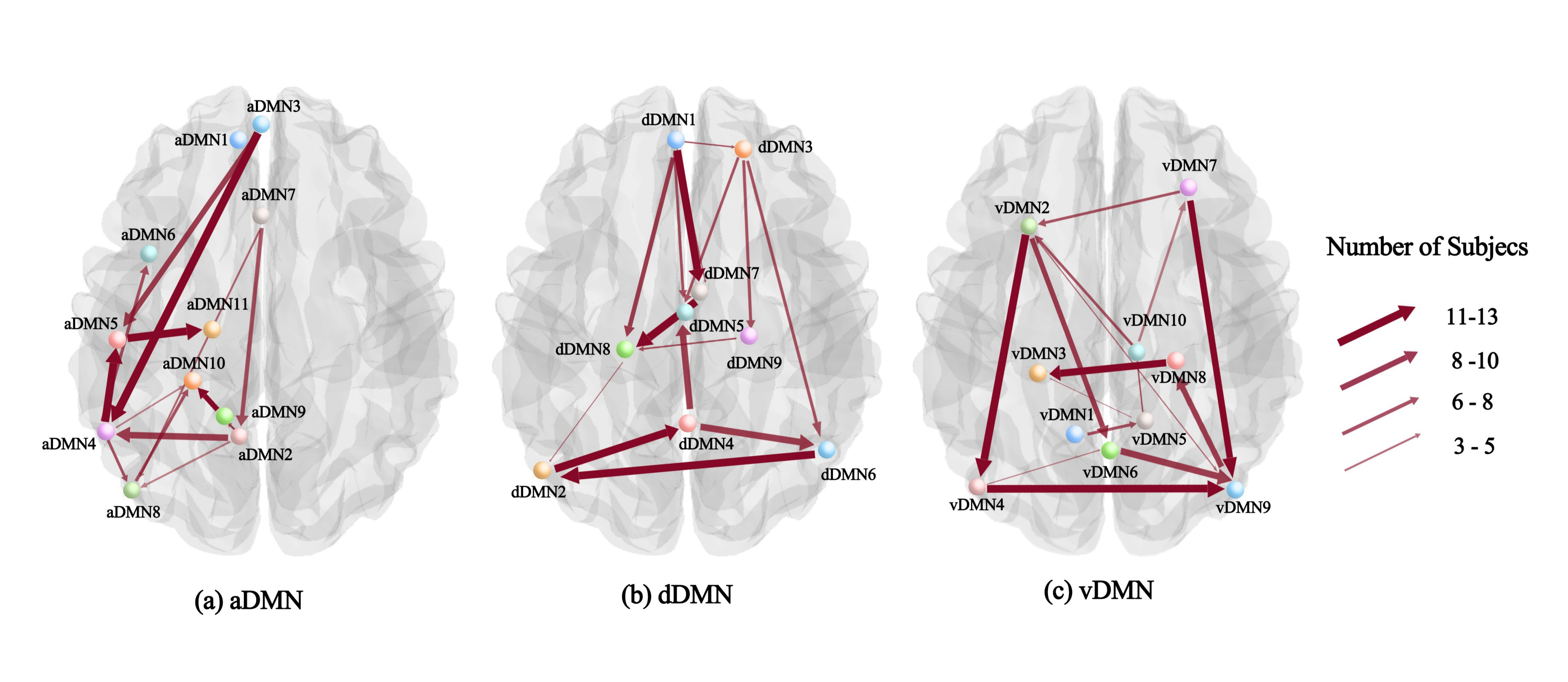}}
  \caption{The network connection counts in three DMN networks of MDD patients.}
  \label{fig:subjectnumber}
\end{figure}

Figure \ref{fig:subjectnumber} depicts the network connection counts (the number of subjects with the same connection) of MDD patients in three DMN networks. ROI connections with high connection counts are generally considered universal connections among MDD patients. Many such ROI connections have been noted as neurophysiological diagnostic biomarkers of MDD. From Figure \ref{fig:subjectnumber}, we can find some important functional connections with high connection counts: aDMN3 (dorsal medial prefrontal cortex) $\rightarrow$ aDMN4 (temporal-parietal junction), aDMN4 (temporal-parietal junction) $\rightarrow$ aDMN5 (lateral temporal cortex), aDMN3 (dorsal medial prefrontal cortex) $\rightarrow$ aDMN5 (lateral temporal cortex ),  aDMN4 (temporal-parietal junction) $\rightarrow$ aDMN10 (parahippocampal cortex), and aDMN9 (retrosplenial cortex) $\rightarrow$ aDMN2 (posterior cingulate cortex). These functional connections are also consistent with the MDD imaging biomarkers identified in the existing literature for distinguishing MDD from HC \citep{khundakar2009morphometric,andrews2010functional,zhao2014brain}.

\begin{table}
\centering
\caption{Two-sample proportion tests for published differential connections.}
\begin{tabular}{clcccc}
\toprule
Template & \multicolumn{1}{c}{Connection} & MTGBN & SIG & GC & DCM \\ \midrule
\multirow{5}{*}{aDMN} & aDMN3 $\rightarrow$ aDMN4 & \textbf{0.0031} & 0.0339 & 0.0127 & 0.2280 \\
& aDMN4 $\rightarrow$ aDMN5 & \textbf{0.0155} & 0.2193 & 0.1367 & 0.3452 \\
& aDMN3 $\rightarrow$ aDMN5 & \textbf{0.0339} & 0.0716 & 0.0607 & 0.1345 \\
& aDMN4 $\rightarrow$ aDMN10 & \textbf{0.0708} & 0.2714 & 0.1279 & 0.2280 \\
& aDMN9 $\rightarrow$ aDMN2 & \textbf{0.0708} & 0.4721 & 0.3563 & 0.3575 \\ \midrule
\multirow{4}{*}{dDMN} & dDMN1 $\rightarrow$ dDMN7 & \textbf{0.0014} & 0.0680 & 0.2321 & 0.1345 \\
& dDMN2 $\rightarrow$ dDMN4 & \textbf{0.0127} & 0.1279 & 0.2321 & 0.1345 \\
& dDMN7 $\rightarrow$ dDMN8 & \textbf{0.0493} & 0.3452 & 0.2045 & 0.3563 \\
& dDMN6 $\rightarrow$ dDMN2 & \textbf{0.1345} & 0.6437 & 0.3524 & 0.6437 \\ \midrule
\multirow{4}{*}{vDMN} & vDMN2 $\rightarrow$ vDMN4 & \textbf{0.0127} & 0.1279 & 0.0327 & 0.3452 \\
& vDMN9 $\rightarrow$ vDMN4 & \textbf{0.0607} & 0.1367 & 0.1367 & 0.6548 \\
& vDMN7 $\rightarrow$ vDMN9 & \textbf{0.1345} & 0.6437 & 0.2280 & 0.7807 \\
& vDMN8 $\rightarrow$ vDMN3 & \textbf{0.1345} & 0.7807 & 0.3563 & 0.6437 \\ \bottomrule
\end{tabular}
\label{tab:pvalue}
\end{table}

To further test whether MTGBN is able to learn the difference between MDD patients and HC participants, we do a two-sample proportion test for each published differential connection, and report the $p$-values in Table \ref{tab:pvalue}. For example, in aDMN there are five functional connections that are known to effectively distinguish MDD from HC (the ``Connection'' column in Table \ref{tab:pvalue}), and a small $p$-value indicates that MTGBN is more likely to identify such a connection in the MDD group than in the HC group. For comparison, the testing results for SIG, GC, and DCM are also included. To ensure the consistency of measurement standards, we control these methods to learn the same density of the network structure.

From Table \ref{tab:pvalue} we can find that overall MTGBN produces the most significant results, whereas in some cases, for example aDMN4 $\rightarrow$ aDMN5, all other methods have insignificant outcomes even at the $\alpha=0.1$ level. This finding validates that MTGBN is good at detecting correlation structures that reflect the actual biological functions.

As an interpretation of the results in Table \ref{tab:pvalue}, we find that within the aDMN template, MDD shows significantly increased connectivity for aDMN3 $\rightarrow$ aDMN4, aDMN4 $\rightarrow$ aDMN5, and aDMN3 $\rightarrow$ aDMN5 compared to HC. According to the definition of DMN subsystem \citep{andrews2010functional}, these are within-system connections in the dorsal medial prefrontal cortex (dMPFC) subsystem, which is considered to play more social reflective roles \citep{andrews2010functional,andrews2014default}, and is related to various cognitive functions, evaluation and expression of negative emotions \citep{etkin2011emotional}. These observations may reflect the resting-state hyperconnectivity in the dMPFC subsystem in MDD patients. The neuropathology studies in MDD also report abnormalities in the dMPFC subsystem \citep{khundakar2009morphometric,zhao2014brain}.

% Furthermore, the inter-system connectivity between the dMPFC subsystem and MTL subsystem increases compared with HC participants. The increase connectivity between subsystems indicates that MDD patients have stronger self-related neural activity, which may be related to the self-abnormalities in MDD \citep{zhu2017rumination}.

% We also compare the performance differences between the proposed MTGBN method and other existing methods (GC and DCM). To ensure the consistency of measurement standards, we control various methods to learn the same density of the network structure. We compare the performance of different methods in identifying important functional connection. The results in Table \ref{tab:pvalue} show that  MTGBN is better than GC and DCM methods in identifying substantial connections. When the GC and DCM methods cannot significantly distinguish the difference between MDD patients and HC participants, MTGBN is still effective.

\section{Conclusion}
\label{sec:conclusion}
Learning Bayesian network structures has received lots of attention, but also encounters significant challenges with insufficient data. This study proposes a multitask learning method to estimate multiple networks simultaneously from related observations, which effectively compensates for the small sample sizes. We achieve this goal by introducing a shared prior covariance matrix among all tasks and utilizing the hierarchical model structure for estimation. Notably, we develop a tractable computing method for the structure learning of networks, powered by analytic likelihood and gradient evaluations that lead to efficient HMC sampling. Extensive and systematic numerical experiments validate the excellent performance of the proposed method, and we highlight its usefulness in analyzing rs-fMRI data and facilitating the study of the major depressive disorder.
\newpage

%%%%%%%%%%%%%%%%%%%%%%%%%%%%%%%%%%%%%%%%%%%%%%
%% Example with multiple Appendixes:        %%
%%%%%%%%%%%%%%%%%%%%%%%%%%%%%%%%%%%%%%%%%%%%%%
\begin{appendix}
\section{Proof of Theorem \ref{Theorem1}}\label{Appendix a}
Here we provide the detailed proof of Theorem \ref{Theorem1}. The conditional distribution $\mathbb{P}\left(B_{i} \mid \Sigma_{h}\right)$ can be expressed as following formula
\[
\mathbb{P}\left(B_{i} \mid \Sigma_{h}\right)=\int \mathbb{P}\left(B_{i} \mid \Sigma_{i}\right) \mathbb{P}\left(\Sigma_{i} \mid \Sigma_{h}\right) d \Sigma_{i},
\]
where the conditional distribution $\mathbb{P}\left(\Sigma_{i} \mid \Sigma_{h}\right)$ is the density function of an inverse Wishart distribution $IW\left(\Sigma_{h}, \nu_{0}\right)$. Therefore, the most challenging part in deriving Theorem \ref{Theorem1} is to measure the density of a sparse BN $B_{i}$ given a covariance matrix $\Sigma_{i}$, \emph{i.e.}, $\mathbb{P}\left(B_{i} \mid \Sigma_{i}\right)$.

Fortunately, \cite{dawid1993hyper} constructs the HIW distribution to address this issue in the context of decomposable undirected graph. Given a decomposable undirected graph $\widetilde{G}_i$ (consider $\widetilde{G}_i$ that is the corresponding undirected graph of $B_i$), the role of HIW distribution is to limit the parameter space of inverse Wishart distribution according to this graph. In the context of undirected graph, we can regard $\mathbb{P}\left(\widetilde{G}_{i} \mid \Sigma_{i}\right) \mathbb{P}\left(\Sigma_{i} \mid \Sigma_{h}\right)$ as a HIW distribution (denote as $HIW(\Sigma_{i}, \Sigma_{h},\widetilde{G}_{i})$). The role of $\mathbb{P}\left(\widetilde{G}_{i} \mid \Sigma_{i}\right)$ here is to limit the parameter space of $\mathbb{P}\left(\Sigma_{i} \mid \Sigma_{h}\right)$ in a subspace $\Psi_{i}$ in which $\Sigma_{i}$ is consistent with $\widetilde{G}_{i}$. Therefore, the HIW distribution $HIW(\Sigma_{i}, \Sigma_{h},\widetilde{G}_{i})$  can be written as follows:
\[
HIW(\Sigma_{i}, \Sigma_{h},\widetilde{G}_{i})=\mathbb{P}\left(\widetilde{G}_{i} \mid \Sigma_{i}\right) \mathbb{P}\left(\Sigma_{i} \mid \Sigma_{h}\right)=\mathbb{P}\left(\Sigma_{i} \mid \Sigma_{h}\right), \Sigma_{i}\in\Psi_{i},
\]
where $\Psi_{i}$ is the subspace of $\Sigma_{i}$. Let $\Omega_{i}=\Sigma_{i}^{-1}$, $\Psi_{i}$ be the set of $p \times p$ positive definite symmetric matrices with $\Omega_{ij}=0$ if and only if $(i,j)\notin \widetilde{G}_{i}$. Therefore, we further derive the following expression:
\begin{equation}
\begin{small}
\mathbb{P}\left(\widetilde{G}_{i} \mid \Sigma_{h}\right)=\int_{\Psi_{i}}  \mathbb{P}\left(\Sigma_{i} \mid \Sigma_{h}\right) d \Sigma_{i},
\quad \Sigma_{i}\in\Psi_{i}.\label{eq:Pbsigma}
\end{small}
\end{equation}

In fact, equation \eqref{eq:Pbsigma} tells the truth that after marginalizing out $\Sigma_{i}$ from $HIW(\Sigma_{i}, \Sigma_{h},\widetilde{G}_{i})$, we can obtain $\mathbb{P}\left(\widetilde{G}_{i} \mid \Sigma_{h}\right)$. Actually, $\mathbb{P}\left(\widetilde{G}_{i} \mid \Sigma_{h}\right)$ is the normalizing constant of $HIW(\Sigma_{i}, \Sigma_{h},\widetilde{G}_{i})$, denote as $h( \Sigma_{h},\widetilde{G}_{i})$, which can be derived to following equation:
\begin{equation}
\mathbb{P}\left(\widetilde{G}_{i} \mid \Sigma^{h}\right)=h( \Sigma_{h},\widetilde{G}_{i})=\frac{\prod_{C \in \mathcal{C}}\left|2^{-1} \Sigma_{h C C}\right|^{\left(\nu_{0}+|C|-1\right) / 2} \Gamma_{|C|}^{-1}\left(\frac{\nu_{0}+|C|-1}{2}\right)}{\prod_{S \in \mathcal{S}}\left|2^{-1} \Sigma_{h S S}\right|^{\left(\nu_{0}+|S|-1\right) / 2} \Gamma_{|S|}^{-1}\left(\frac{\nu_{0}+|S|-1}{2}\right)},
\label{eq:h}
\end{equation}
where $\Gamma(\cdot)$ denotes  Gamma function, $\mathcal{C}$ and $\mathcal{S}$ are the cliques and separators of the decomposable undirected graph, $\Sigma_{h C C}$ and $\Sigma_{h S S}$ denote the submatrix of $\Sigma_{h}$ according to graph decomposation, $|C|$ and $|S|$ denote the cardinality of the clique $C$ and separator $S$ respectively.

Fortunately, a BN $B_{i}$ has an unique undirected graph $\widetilde{G}_{i}$ corresponding to it, thus we have the following relationship
\[
\mathbb{P}\left(B_{i} \mid \Sigma_{h}\right) =\mathbb{P}\left(\widetilde{G}_{i} \mid \Sigma_{h}\right)= h( \Sigma_{h},\widetilde{G}_{i}).
\]

\section{Proof of Theorem \ref{Theorem2}} \label{Appendix b} Here we provide the detailed proof of Theorem \ref{Theorem2}. Let $\boldsymbol{X}$ be a random vector of $p$ continuous variables $(X_{1}, \ldots, X_{p})^\top$, and $D=\left\{\bx_{1}, \ldots, \bx_{n}\right\}$ be $n$ observations of $\boldsymbol{X}$. Assume $\boldsymbol{X}$ follows a multivariate normal distribution, and $B$ is the GBN associated with $\boldsymbol{X}$. Then, \cite{geiger1994learning} shows that the conditional distribution of $D$ given the network structure $B$ can be factorized as 
\begin{equation}
\mathbb{P}\left(D \mid B\right)=\prod_{k=1}^{p} \frac{\mathbb{P}\left(D^{\left\{X_{k}\right\} \cup \Pi_{k}}  \right)}{\mathbb{P}\left(D^{\Pi_{k}} \right)},
\label{eq:Pdb}
\end{equation}
where $D^{\left\{X_{k}\right\} \cup \Pi_{k}}$ is the dataset $D$ restricted to the variable $X_{k}$ and its parents $\Pi_{k}$. Each term in equation \eqref{eq:Pdb} follows a multivariate normal distribution.

Similar to equation \eqref{eq:Pdb}, we can derive  $\mathbb{P}\left(D_{i} \mid B_{i}, \Sigma_{h}\right)$ by encoding the BN structure $ B_{i}$ into $\mathbb{P}(D_{i}|\Sigma_{h})$, note that:
\[
\mathbb{P}\left(D_{i} \mid B_{i}, \Sigma_{h}\right)=\prod_{k=1}^{p} \frac{\mathbb{P}\left(D_{i}^{\left\{X_{k}\right\} \cup \Pi_{k}} \mid \Sigma_{h}\right)}{\mathbb{P}\left(D_{i}^{\Pi_{k}} \mid \Sigma_{h}\right)}.
\]

Therefore, the main fact we need to do is to derive $\mathbb{P}(D_{i}|\Sigma_{h})$. From equation \eqref{eq:Psigma} and \eqref{eq:Pdsigmai}, we have
\begin{align}\nonumber
\mathbb{P}(D_{i},\Sigma_{i}|\Sigma_{h})\nonumber&=\mathbb{P}(D_{i}|\Sigma_{i})\mathbb{P}(\Sigma_{i}|\Sigma_{h})\nonumber\\
&=\frac{(2\pi)^{-n_{i}p/2}|\Sigma_{h}|^{\nu_{0}/2}}{2^{\nu_{0}p/2}\Gamma_{p}(\nu_{0}/2)}|\Sigma_{i}|^{-(\nu_{0}+p+n_{i}+1)/2}\exp\left\{ -\frac{1}{2}\mathrm{tr}\left[\left(\Sigma_{h}+n_{i}S_{i}\right)\Sigma_{i}^{-1}\right]\right\} .
\label{eq:Pdsigma2}
\end{align}

Moreover, since $\mathbb{P}(D_{i}|\Sigma_{i}){\sim} \mathcal{N}(0,\Sigma_{i})$
and $\mathbb{P}(\Sigma_{i}|\Sigma_{h})\sim IW(\Sigma_{h},\nu_{0})$. Due to the conjugacy, we have $\mathbb{P}(\Sigma_{i}|D_{i},\Sigma_{h})\sim IW(\Sigma_{h}+n_{i}S_{i},\nu_{0}+n_{i})$, which can be written as
\[
\mathbb{P}(\Sigma_{i}|D_{i},\Sigma_{h})=\frac{|\Sigma_{h}+n_{i}S_{i}|^{(\nu_{0}+n_{i})/2}}{2^{(\nu_{0}+n_{i})p/2}\Gamma_{p}\left(\frac{\nu_{0}+n_{i}}{2}\right)}|\Sigma_{i}|^{-(\nu_{0}+p+n_{i}+1)/2}\exp\left\{ -\frac{1}{2}\mathrm{tr}\left[\left(\Sigma_{h}+n_{i}S_{i}\right)\Sigma_{i}^{-1}\right]\right\}.
\]
and as a result, we obtain
\[
\mathbb{P}(D_{i}|\Sigma_{h})=\frac{\mathbb{P}(D_{i},\Sigma_{i}|\Sigma_{h})}{\mathbb{P}(\Sigma_{i}|D_{i},\Sigma_{h})}=\frac{|\Sigma_{h}|^{\nu_{0}/2}\Gamma_{p}\left(\frac{\nu_{0}+n_{i}}{2}\right)}{\pi^{n_{i}p/2}|\Sigma_{h}+n_{i}S_{i}|^{(\nu_{0}+n_{i})/2}\Gamma_{p}\left(\frac{\nu_{0}}{2}\right)}\label{eq:data_given_sigma_h}.
\]

\section{Proof of Theorem~\ref{Theorem3}} \label{Appendix c}
In this section we provide the detailed proof of Theorem \ref{Theorem3}. To derive $\mathbb{P}(\Sigma_{h}|\mathcal{B},\mathcal{D})$, where $\mathcal{B}=\left\{B_{i}\right\}_{i=1}^{m}$, note that
\[
\mathbb{P}(\Sigma_{h}|\mathcal{B},\mathcal{D})\propto\mathbb{P}(\Sigma_{h},\mathcal{B},\mathcal{D})\propto\mathbb{P}(\mathcal{B},\mathcal{D}|\Sigma_{h})\mathbb{P}(\Sigma_{h})\propto\mathbb{P}(\mathcal{B},\mathcal{D}|\Sigma_{h}).
\]

By conditioning on $\{\Sigma_{i}\}_{i=1}^{m}$, we obtain
\begin{align}
&\mathbb{P}(\mathcal{B},\mathcal{D}|\Sigma_{h})\nonumber\\
&  =\int\mathbb{P}(\mathcal{B}|\{\Sigma_{i}\}_{i=1}^{m})\mathbb{P}(\mathcal{D}|\{\Sigma_{i}\}_{i=1}^{m})\mathbb{P}(\{\Sigma_{i}\}_{i=1}^{m}|\Sigma_{h})d\Sigma_{1}\cdots d\Sigma_{m}\nonumber\\
 & =\prod_{i=1}^{m}\int \mathbb{P}(B_{i}|\Sigma_{i})\mathbb{P}(D_{i}|\Sigma_{i})\mathbb{P}(\Sigma_{i}|\Sigma_{h})d\Sigma_{i}.
\label{eq:Pbdsigma}
\end{align}

The challenging of derivation in equation \eqref{eq:Pbdsigma} is how to integrate $\Sigma_{i}$ and generate $B_{i}$ and $D_{i}$ from $\Sigma_{h}$. Fortunately, Theorem \ref{Theorem1} provides us a bridge so that we can derive $\mathbb{P}(\mathcal{B},\mathcal{D}|\Sigma_{h})$ without actually integrating $\Sigma_{i}$. Thus, we can re-write equation \eqref{eq:Pbdsigma} as :
\[
\mathbb{P}(\mathcal{B},\mathcal{D}|\Sigma_{h})=\prod_{i=1}^{m}\int_{\Psi_{i}}\mathbb{P}(D_{i}|\Sigma_{i})\mathbb{P}(\Sigma_{i}|\Sigma_{h})d\Sigma_{i},
\quad \Sigma_{i}\in\Psi_{i}.
\]

In equation \eqref{eq:Pdsigma2}, we have already shown that

\[
\mathbb{P}(D_{i}|\Sigma_{i})\mathbb{P}(\Sigma_{i}|\Sigma_{h})=\frac{(2\pi)^{-n_{i}p/2}|\Sigma_{h}|^{\nu_{0}/2}}{2^{\nu_{0}p/2}\Gamma_{p}(\nu_{0}/2)}|\Sigma_{i}|^{-(\nu_{0}+p+n_{i}+1)/2}\exp\left\{ -\frac{1}{2}\mathrm{tr}\left[\left(\Sigma_{h}+n_{i}S_{i}\right)\Sigma_{i}^{-1}\right]\right\}.
\]
So we need to compute the integral
\begin{equation}
\int_{\Psi_{i}}|\Sigma_{i}|^{-(\nu_{0}+p+n_{i}+1)/2}\exp\left\{ -\frac{1}{2}\mathrm{tr}\left[\left(\Sigma_{h}+n_{i}S_{i}\right)\Sigma_{i}^{-1}\right]\right\} d\Sigma_{i}.
\label{eq:integral}
\end{equation}
Recall that $\mathbb{P}(\Sigma_{i}|D_{i},\Sigma_{h})\sim IW(\Sigma_{h}+n_{i}S_{i},\nu_{0}+n_{i})$, so
\[
\mathbb{P}(\Sigma_{i}|D_{i},\Sigma_{h})=\frac{|\Sigma_{h}+n_{i}S_{i}|^{(\nu_{0}+n_{i})/2}}{2^{(\nu_{0}+n_{i})p/2}\Gamma_{p}\left(\frac{\nu_{0}+n_{i}}{2}\right)}|\Sigma_{i}|^{-(\nu_{0}+p+n_{i}+1)/2}\exp\left\{ -\frac{1}{2}\mathrm{tr}\left[\left(\Sigma_{h}+n_{i}S_{i}\right)\Sigma_{i}^{-1}\right]\right\},
\]
which is proportional to the integral in equation \eqref{eq:integral}. Therefore,
\begin{align}
&\int_{\Psi_{i}}|\Sigma_{i}|^{-(\nu_{0}+p+n_{i}+1)/2}\exp\left\{ -\frac{1}{2}\mathrm{tr}\left[\left(\Sigma_{h}+n_{i}S_{i}\right)\Sigma_{i}^{-1}\right]\right\} d\Sigma_{i}\nonumber \\
= & \frac{2^{(\nu_{0}+n_{i})p/2}\Gamma_{p}\left(\frac{\nu_{0}+n_{i}}{2}\right)}{|\Sigma_{h}+n_{i}S_{i}|^{(\nu_{0}+n_{i})/2}}\int_{\Psi_{i}}\mathbb{P}(\Sigma_{i}|\mathcal{D}_{i},\Sigma_{h})d\Sigma_{i}.\label{eq:integral2}
\end{align}

Comparing $\mathbb{P}(\Sigma_{i}|\Sigma_{h})$ in the equation \eqref{eq:Pbsigma} and $\mathbb{P}(\Sigma_{i}|\mathcal{D}_{i},\Sigma_{h})d\Sigma_{i}$ in equation \eqref{eq:integral2}, we find that they only differ in the parameters of the inverse Wishart distribution. Then we conclude that $\int_{\Psi_{i}}\mathbb{P}(\Sigma_{i}|D_{i},\Sigma_{h})d\Sigma_{i}$ has a similar closed-form expression to \eqref{eq:h}, given by
\[
\int_{\Psi_{i}}\mathbb{P}(\Sigma_{i}|D_{i},\Sigma_{h})\mathrm{d}\Sigma_{i}=h(\Sigma_{h}+n_{i}S_{i}, \widetilde{G}_{i}).
\]
As a result,
\[
\begin{aligned}
&\int_{\Psi_{i}}|\Sigma_{i}|^{-(\nu_{0}+p+n_{i}+1)/2}\exp\left\{ -\frac{1}{2}\mathrm{tr}\left[\left(\Sigma_{h}+n_{i}S_{i}\right)\Sigma_{i}^{-1}\right]\right\} d\Sigma_{i}\nonumber \\
= & \frac{2^{(\nu_{0}+n_{i})p/2}\Gamma_{p}\left(\frac{\nu_{0}+n_{i}}{2}\right)}{|\Sigma_{h}+n_{i}S_{i}|^{(\nu_{0}+n_{i})/2}}\int_{\Psi_{i}}\mathbb{P}(\Sigma_{i}|\mathcal{D}_{i},\Sigma_{h},\nu_{0})d\Sigma_{i},
\end{aligned}
\]
and
\begin{small}
\[
\mathbb{P}(\Sigma_{h}|\mathcal{B},\mathcal{D})=\prod_{i=1}^{m}\frac{|\Sigma_{h}|^{v_{0}/2}\cdot h(\Sigma_{h}+n_{i}S_{i}, \widetilde{G}_{i})}{|\Sigma_{h}+n_{i}S_{i}|^{(\nu_{0}+n_{i})/2}}.
\]
\end{small}

\section{Details of HMC Algorithm \label{Appendix d}}
In this section we continue the discussion on the HMC algorithm in Step 5 of Algorithm \ref{mcem}. The trickiest thing about sampling $\Sigma_{h}$ from $\mathbb{P}\left(\Sigma_{h}\mid \mathcal{B}, \mathcal{D}\right)$ is that $\Sigma_{h}$ is constrained to the positive semidefinite cone. Next, we discuss how to transform $\Sigma_{h}$  to a unconstrained  lower-triangular matrix $V$ (\textbf{Transform}) and inverse-transform from $V$ to $\Sigma_{h}$ (\textbf{Inverse-Transform}).

\textbf{Transform}: For the positive semidefinite matrix $\Sigma_{h}$, there is a unique lower-triangular matrix $L=chol(\Sigma_{h})$ with positive diagonal entries, call a Cholesky factor, such that:
\[
\Sigma_{h}=LL^{\mathrm{T}}.
\]

The off-diagonal entries of the Cholesky factor $L$ are unconstrained, but the diagonal entries $l_{ii}$ must be positive. To complete the transform, the diagonal is log-transformed to produce a fully unconstrained lower-triangular matrix $V$ defined by
\[
v_{ij}=\left\{\begin{array}{cl}
0 & \text { if } i<j, \\
\log l_{ii} & \text { if } i=j,\\
l_{ij} & \text { if } i>j .
\end{array}\right.
\]

\textbf{Inverse-Transform}: The inverse-transform reverses the two steps of the transform. Given an unconstrained lower-triangular matrix $V$, the first step is to recover the intermediate matrix $L$ by reversing the log trandfrom,
\[
l_{ij}=\left\{\begin{array}{cl}
0 & \text { if } i<j, \\
\exp v_{ii} & \text { if } i=j, \\
v_{ij} & \text { if } i>j .
\end{array}\right.
\]

In Section \ref{sec:computing}, we derived the expression of $\mathbb{P}\left(V\mid \mathcal{B}, \mathcal{D}\right)$. The core of HMC algorithm is the calculation of logarithmic density function and its gradient. Thus, next we will compute the logarithm of $\mathbb{P}\left(V\mid \mathcal{B}, \mathcal{D}\right)$ and its gradient. The logarithm of $\mathbb{P}\left(V\mid \mathcal{B}, \mathcal{D}\right)$, which denotes as $\log\{\mathbb{P}(V|\mathcal{B},\mathcal{D})\}$ that can be expressed as follows:
\begin{align}\nonumber
\log\{\mathbb{P}(V|\mathcal{B},\mathcal{D})\}& =\sum_{i=1}^{m}\log h(\Sigma_{h}+n_{i}S_{i}, B_{i})-\sum_{i=1}^{m}\frac{\nu_{0}+n_{i}}{2}\log|\Sigma_{h}+n_{i}S_{i}|\nonumber \\
 & \quad+\sum_{i=1}^{p}(m\nu_{0}+p-i+2)v_{ii}+C.
\label{eq:logV}
\end{align}

After deriving $\log\{\mathbb{P}(V|\mathcal{B},\mathcal{D})\}$ we also need to derive the gradient of $\log\{\mathbb{P}(V|\mathcal{B},\mathcal{D})\}$ with respect to $V$, which denote as $ \nabla_{V}\log\{\mathbb{P}(V|\mathcal{B},\mathcal{D})\}$.
The expression of $\log\{\mathbb{P}(V|\mathcal{B},\mathcal{D})\}$ mainly includes three parts, we will compute the gradient for each term separately.

The first term consists of functions of the form $g(\Sigma_{h})=\log|(\Sigma_{h}+A)_{CC}|$,
where $C$ is an index set. Let $I_{C}$ be a matrix constructed by
selecting rows of the identity matrix indexed by $C$, and then $\Sigma_{hCC}=I_{C}\Sigma_{h}I_{C}^{\mathrm{T}}=I_{C}LL^{\mathrm{T}}I_{C}^{\mathrm{T}}$.
It is well known that $\partial g/\partial\Sigma_{hCC}=\{(\Sigma_{h}+A)_{CC}^{-1}\}^{\mathrm{T}}=(\Sigma_{h}+A)_{CC}^{-1}$,
so we can write $\mathrm{d}g=\mathrm{tr}\{(\Sigma_{h}+A)_{CC}^{-1}(\mathrm{d}\Sigma_{hCC})\}$.
Since

\[
\mathrm{d}\Sigma_{hCC}=I_{C}(\mathrm{d}\Sigma_{h})I_{C}^{\mathrm{T}}=I_{C}L(\mathrm{d}L)^{\mathrm{T}}I_{C}^{\mathrm{T}}+I_{C}(\mathrm{d}L)L^{\mathrm{T}}I_{C}^{\mathrm{T}}.
\]
Then we have
\[
\begin{aligned}
\mathrm{d}g & =\mathrm{tr}\{(\mathrm{d}L)^{\mathrm{T}}I_{C}^{\mathrm{T}}(\Sigma_{h}+A)_{CC}^{-1}I_{C}L\}+\mathrm{tr}\{L^{\mathrm{T}}I_{C}^{\mathrm{T}}(\Sigma^{h}+A)_{CC}^{-1}I_{C}(\mathrm{d}L)\}\\
 & =2\mathrm{tr}\{L^{\mathrm{T}}I_{C}^{\mathrm{T}}(\Sigma^{h}+A)_{CC}^{-1}I_{C}(\mathrm{d}L)\}.
\end{aligned}
\]

Recall that $L$ is lower-triangular, so we get $\partial g/\partial L=2\mathbf{lower}\{I_{C}^{\mathrm{T}}(\Sigma_{h}+A)_{CC}^{-1}I_{C}L\}$, where $\mathbf{lower}\{A\}$ stands for the lower-triangular part
of a matrix $A$. Finally, it is easy to see that
\[
\frac{\partial g}{\partial v_{ij}}=\begin{cases}
\frac{\partial g}{\partial l_{ii}}\cdot l_{ii}, & i=j,\\
\frac{\partial g}{\partial l_{ij}}, & i>j.
\end{cases}
\]

The second term involves $f(\Sigma_{h})=\log|\Sigma_{h}+A|$, where
$A$ is a positive definite matrix. Similar to $\partial g/\partial\Sigma_{hCC}$, we have $\partial f/\partial\Sigma_{h}=(\Sigma_{h}+A)^{-1}$, and
$\mathrm{d}f=\mathrm{tr}\{(\Sigma_{h}+A)^{-1}(\mathrm{d}\Sigma_{h})\}$. Again using the relation $\mathrm{d}\Sigma_{h}=L(\mathrm{d}L)^{\mathrm{T}}+(\mathrm{d}L)L^{\mathrm{T}}$, we obtain
\[
\begin{aligned}
\mathrm{d}f & =\mathrm{tr}\{(\Sigma_{h}+A)^{-1}(L(\mathrm{d}L)^{\mathrm{T}}+(\mathrm{d}L)L^{\mathrm{T}})\}\\
 & =\mathrm{tr}\{(\mathrm{d}L)^{\mathrm{T}}(\Sigma_{h}+A)^{-1}L\}+\mathrm{tr}\{L^{\mathrm{T}}(\Sigma_{h}+A)^{-1}(\mathrm{d}L)\}\\
 & =2\mathrm{tr}\{L^{\mathrm{T}}(\Sigma_{h}+A)^{-1}(\mathrm{d}L)\}.
\end{aligned}
\]
So $\partial f/\partial L=2\mathbf{lower}\{(\Sigma_{h}+A)^{-1}L\}$.
Similar to $\partial g/\partial v_{ij}$ we have
of a matrix $A$. Finally, it is easy to see that
\[
\frac{\partial f}{\partial v_{ij}}=\begin{cases}
\frac{\partial f}{\partial l_{ii}}\cdot l_{ii}, & i=j,\\
\frac{\partial f}{\partial l_{ij}}, & i>j.
\end{cases}
\]
Thus $\nabla_{V} \log \mathbb{P}(V|\mathcal{B},\mathcal{D})$ can be expressed as follows:
\begin{equation}
\begin{aligned}
	\mathbf{diag}\{\nabla_{V} \log \mathbb{P}(V|\mathcal{B},\mathcal{D})\}&=\mathbf{diag}\{\widetilde{L}+q(m,\nu_{0},p)\cdot I\} \circ \mathbf{diag}\{L\},\\
	\mathbf{lower}\{\nabla_{V} \log \mathbb{P}(V|\mathcal{B},\mathcal{D})\}&=\mathbf{lower}\{\widetilde{L}\},
\end{aligned}
\end{equation}
where
\begin{align*}
	\widetilde{L}&=g(\nu_{0}+n_{i},\Sigma_{h}+n_{i} S_{i}, \mathcal{B})-f(\nu_{0}+n_{i},\Sigma_{h}+n_{i} S_{i}),\\
	g(\nu,\Sigma,\mathcal{B})&=2\nu\left\{\sum_{C \in \mathcal{C}} I_{C}^{\mathrm{T}}(\Sigma_{CC})^{-1}I_{C}L-\sum_{S \in \mathcal{S}}I_{S}^{\mathrm{T}}(\Sigma_{SS})^{-1}I_{S}L\right\},\\
	f(\nu,\Sigma)&=\nu \cdot \Sigma^{-1}L,\\
	q(m,\nu,p)&=\frac{p(p+2m\nu+3)}{2}.
\end{align*}

According to the above definition, the specific solution process and iterative steps of HMC algorithm are shown in Algorithm \ref{alg:HMC}.
%\begin{figure}[] 
		%\renewcommand{\algorithmicrequire}{\textbf{Input:}}
		%\renewcommand{\algorithmicensure}{\textbf{Output:}}
		%\removelatexerror
		\begin{algorithm}[h]
			\caption{HMC for Sampling $\Sigma_h$}
			\begin{algorithmic}[1]\label{alg:HMC}
				\REQUIRE   Observational datasets $\mathcal{D}$, BN structures $\mathcal{B}$ and initial $\Sigma_{h}^{(0)}$.
				\ENSURE $\{\Sigma_{h}^{(l)}\}_{l=1}^{N}$: $l$ samples of $\Sigma_{h}$.
				\STATE \textbf{Initialization:} The number of iterations $N$, the number of steps $K$, the step length $\epsilon$, and a specified covariance matrix to simulate momentum $M$.
				\STATE $V^{(0)}$ $\leftarrow $ Transform ($\Sigma_{h}^{(0)}$).
				\STATE $\log\{\mathbb{P}(V|\mathcal{B},\mathcal{D})\}$ = equation \eqref{eq:logV}.
				\STATE Calculate $\log\{\mathbb{P}(V^{(0)}|\mathcal{B},\mathcal{D})\}$.
				\FOR{$t=1, \ldots, N$}
					\STATE $\mathbf{p} \leftarrow \mathcal{N} (0, \mathbf{M})$,
					\STATE $V^{(t)} \leftarrow V^{(t-1)}, \tilde{V} \leftarrow V^{(t-1)}, \tilde{\mathbf{p}} \leftarrow \mathbf{p}$,
					\FOR{$i=1, \ldots, K$}
					   \STATE $\tilde{V}, \tilde{\mathbf{p}} \leftarrow$ Leapfrog $(\tilde{V}, \tilde{\mathbf{p}}, \epsilon, \mathbf{M})$.
					\ENDFOR
					\STATE $\alpha \leftarrow \min \left(1, \frac{\exp \left(\log f(\tilde{V})-\frac{1}{2} \tilde{\mathbf{p}}^{T} \mathbf{M}^{-1} \tilde{\mathbf{p}}\right)}{\exp \left(\log f\left(\tilde{V}^{(t-1)}\right)-\frac{1}{2} \mathbf{p}^{T} \mathbf{M}^{-1} \mathbf{p}\right)}\right)$,
					\STATE With probability $\alpha, V^{(t)} \leftarrow \tilde{V} \text { and } \mathbf{p}^{(t)} \leftarrow-\tilde{\mathbf{p}}$,
					\STATE  $\Sigma_{h}^{(t)}$ $\leftarrow$ Inverse-Transform ($V^{(t)}$).
				\ENDFOR
			   	\RETURN $\Sigma_{h}^{(1)}, \ldots, \Sigma_{h}^{(N)}$.
				\STATE \textbf{function} Leapfrog $\left(V^{*}, \mathbf{p}^{*}, \epsilon, \mathbf{M}\right)$
				\STATE \quad $\tilde{\mathbf{p}} \leftarrow \mathbf{p}^{*}+(\epsilon / 2) \nabla_{V} \log f\left(V^{*}\right)$,
				\STATE \quad $\tilde{V} \leftarrow V^{*}+\epsilon \mathbf{M}^{-1} \tilde{\mathbf{p}}$,
				\STATE \quad $\tilde{\mathbf{p}} \leftarrow \tilde{\mathbf{p}}+(\epsilon / 2) \nabla_{V} \log f(\tilde{V})$,
				\STATE \quad \textbf{return} $\tilde{V}, \tilde{\mathbf{p}}$.
				\STATE \textbf{end function}
			\end{algorithmic}
		\end{algorithm}
%\end{figure}

\section{ROI information of three DMN networks}

See Table \ref{tab: roi}.

\begin{table}[]
\centering
\caption{Description of ROI nodes in the three DMN networks and their Montreal Neurological Institute (MNI) coordinates.}
\begin{tabular}{@{}cccccclllllll@{}}
\toprule
Region name & Abbreviation & Broadman Areas & x & y & z   \\ \midrule
\multicolumn{1}{l}{\textbf{aDMN (Andrews-Hanna Default Mode Network)}}    \\
\multicolumn{1}{l}{\textit{\textbf{P (PCC-aMPFC Core)}}}   \\
Anterior medial prefrontal cortex & aDMN1 & 10, 32 & -6 & 52 & -2   \\
Posterior cingulate cortex & aDMN2 & 23, 31 & -8 & -56 & 26  \\
\multicolumn{1}{l}{\textit{\textbf{D (dMPFC Subsystem)}}} & \multicolumn{1}{l}{} & \multicolumn{1}{l}{} & \multicolumn{1}{l}{} & \multicolumn{1}{l}{} & \multicolumn{1}{l}{} \\
Dorsal medial prefrontal cortex & aDMN3 & 9, 32 & 0 & 52 & 26   \\
Temporal parietal junction & aDMN4 & 40, 39 & -54 & -54 & 28   \\
Lateral temporal cortex & aDMN5 & 21, 22 & -60 & -24 & -18   \\
Temporal pole & aDMN6 & 21 & -50 & 14 & -40  \\
\multicolumn{1}{l}{\textit{\textbf{M (MTL Subsystem)}}} & \multicolumn{1}{l}{} & \multicolumn{1}{l}{} & \multicolumn{1}{l}{} & \multicolumn{1}{l}{} & \multicolumn{1}{l}{}  \\
Ventral medial prefrontal cortex & aDMN7 & 11, 24, 25, 32 & 0 & 26 & -18   \\
Posterior inferior parietal lobule & aDMN8 & 39 & -44 & -74 & 32  \\
Retrosplenial cortex & aDMN9 & 29, 30, 19 & -14 & -52 & 8  \\
Parahippocampal cortex & aDMN10 & 20, 36, 19 & -28 & -40 & -12   \\
Hippocampal formation & aDMN11 & 20, 36 & -22 & -20 & -26 &  \\
\multicolumn{1}{l}{\textbf{dDMN (Dorsal Default Mode Network)}} & \multicolumn{1}{l}{} & \multicolumn{1}{l}{} & \multicolumn{1}{l}{} & \multicolumn{1}{l}{} & \multicolumn{1}{l}{} \\
Medical anterior cingulate cortex & dDMN1 & 9, 10, 24, 32, 11 & -3 & 49 & 14  \\
Left angular gyrus & dDMN2 & 39 & -48 & -68 & 35  \\
Right superior frontal gyrus & dDMN3 & 9 & 19 & 38 & 47 \\
Posterior cingulate cortex, precuneus & dDMN4 & 23, 30 & 1 & -53 & 28 \\
Middle cingulate cortex & dDMN5 & 23 & 2 & -15 & 36   \\
Right angular gyrus & dDMN6 & 29 & 56 & -66 & 36 \\
Left and right thalamus & dDMN7 & N/A & 6 & -6 & 2  \\
Left hippocampus & dDMN8 & 20, 36, 30 & -24 & -29 & -13  \\
Right hippocampus & dDMN9 & 20, 36, 30 & 27 & -23 & -17   \\
\multicolumn{1}{l}{\textbf{vDMN (Ventral Default Mode Network)}} & \multicolumn{1}{l}{} & \multicolumn{1}{l}{} & \multicolumn{1}{l}{} & \multicolumn{1}{l}{} & \multicolumn{1}{l}{}  \\
Left posterior cingulate cortex & vDMN1 & 29, 30, 23 & -12 & -58 & 15  \\
Left middle frontal gyrus & vDMN2 & 8, 6 & -24 & 12 & 55  \\
Left parahippocampal  gyrus & vDMN3 & 37, 20 & -28 & -37 & -15  \\
Left middle occipital gyrus & vDMN4 & 19, 39 & -44 & -74 & 30 \\
Right  posterior cingulate cortex & vDMN5 & 30, 23 & 13 & -53 & 14   \\
Precuneus & vDMN6 & 7, 5 & 1 & -57 & 54 \\
Right superior frontal gyrus, & vDMN7 & 9, 8 & 26 & 26 & 45  \\
Right parahippocampal gyrus & vDMN8 & 37, 30 & 28 & -33 & -19  \\
Right angular gyrus & vDMN9 & 39, 19 & 43 & -74 & 32  \\
Right lobule & vDMN10 & N/A & 8 & -26 & 70   \\ \bottomrule
\end{tabular}
\label{tab: roi}
\end{table}

\end{appendix}

\newpage

\bibliographystyle{imsart-nameyear} 
\bibliography{bibliography} 
\end{document}